\documentclass{article} % For LaTeX2e
\usepackage{iclr2022_conference,times}

\usepackage{amsmath,amsfonts,bm}

\def\eqref#1{equation~\ref{#1}}
\def\1{\bm{1}}

\DeclareMathAlphabet{\mathsfit}{\encodingdefault}{\sfdefault}{m}{sl}
\SetMathAlphabet{\mathsfit}{bold}{\encodingdefault}{\sfdefault}{bx}{n}

\usepackage{hyperref}
\usepackage{url}
\usepackage{amsthm}
\usepackage{amsmath}

\usepackage{appendix}
\usepackage{enumitem}
\usepackage{amsfonts}
\usepackage{multirow}
\usepackage{multicol}

\usepackage{color}
\usepackage{algorithm}
\usepackage{algorithmic}

\usepackage{graphicx}
\usepackage{subfig}
\usepackage{hyperref}

\newtheorem{theorem}{Theorem}
\newtheorem{lemma}{Lemma}

\newtheorem{assumption}{Assumption}
\newtheorem{remark}{Remark}

\title{ Bregman Gradient Policy Optimization }

\author{Feihu Huang\thanks{ Feihu and Shangqian contributed equally.
} \ \thanks{ Corresponding Authors.
}, \ Shangqian Gao$^*$, \  Heng Huang$^\dag$  \\
Department of Electrical and Computer Engineering\\
University of Pittsburgh\\
Pittsburgh, PA 15261, USA \\
\texttt{huangfeihu2018@gmail.com, shg84@pitt.edu, heng.huang@pitt.edu} \\
}

\iclrfinalcopy % Uncomment for camera-ready version, but NOT for submission.
\begin{document}

\maketitle

\begin{abstract}
In the paper, we design a novel Bregman gradient policy optimization framework for reinforcement learning
based on Bregman divergences and momentum techniques.
Specifically, we propose a Bregman gradient policy optimization (BGPO) algorithm based on the basic momentum technique and mirror descent iteration.
Meanwhile, we further propose an accelerated Bregman gradient policy optimization (VR-BGPO) algorithm based on the variance reduced technique.
Moreover, we provide a convergence analysis framework for our Bregman gradient policy optimization under the nonconvex setting.
We prove that our BGPO achieves a  sample complexity of $O(\epsilon^{-4})$ for finding $\epsilon$-stationary policy
only requiring one trajectory at each iteration,
and our VR-BGPO reaches the best known sample complexity of $O(\epsilon^{-3})$,
which also only requires one trajectory at each iteration.
In particular, by using different Bregman divergences, our BGPO framework unifies many existing policy optimization algorithms such as
the existing (variance reduced) policy gradient algorithms such as natural policy gradient algorithm.
Extensive experimental results on multiple reinforcement learning tasks demonstrate the efficiency of our new algorithms.
\end{abstract}
\vspace*{-8pt}
\section{ Introduction }
\vspace*{-8pt}
Policy Gradient (PG) methods are a class of popular policy optimization methods for Reinforcement Learning (RL),
and have achieved significant successes in many challenging applications \citep{li2017deep}
such as robot manipulation \citep{deisenroth2013survey}, the Go game \citep{silver2017mastering} and autonomous driving \citep{shalev2016safe}.
In general, PG methods directly search for the optimal policy by maximizing the expected total reward of Markov
Decision Processes (MDPs) involved in RL, where an agent takes action dictated by a policy in an unknown
dynamic environment over a sequence of time steps. Since the PGs are generally estimated by Monte-Carlo sampling,
such vanilla PG methods usually suffer from very high variances resulted in slow convergence rate and destabilization.
Thus, recently many fast PG methods have been proposed to reduce variances in vanilla stochastic PGs.
For example, \cite{sutton2000policy} introduced a baseline to reduce variances of the stochastic PG.
\cite{konda2000actor} proposed an efficient actor-critic algorithm by estimating the value function to reduce effects of large variances. \citep{schulman2015high} proposed the generalized advantage estimation (GAE) to control
both the bias and variance in policy gradient. More recently, some faster variance-reduced PG methods \citep{papini2018stochastic,xu2019improved,shen2019hessian,liu2020improved,huang2020momentum} have been developed based on the
variance-reduction techniques in stochastic optimization.

Alternatively, some successful PG algorithms \citep{schulman2015trust,schulman2017proximal} improve convergence rate
and robustness of vanilla PG methods by using some penalties such as
Kullback-Leibler (KL) divergence penalty. For example, trust-region policy optimization (TRPO) \citep{schulman2015trust} ensures that the new selected policy is near to the old one by using KL-divergence constraint,
while proximal policy optimization (PPO) \citep{schulman2017proximal} clips the weighted likelihood ratio to implicitly reach this goal.
Subsequently, \citet{shani2020adaptive} have analyzed the global convergence properties of TRPO in tabular RL based on
the convex mirror descent algorithm.
\citet{liu2019neural} have also studied the global convergence properties of PPO and TRPO equipped with overparametrized neural networks
based on mirror descent iterations. At the same time, \citet{yang2019policy} tried to propose the PG methods
based on the mirror descent algorithm.
More recently, mirror descent policy optimization (MDPO) \citep{tomar2020mirror} iteratively updates the policy
beyond the tabular RL by approximately solving a trust region problem based on
convex mirror descent algorithm. In addition, \cite{agarwal2019theory,cen2020fast} have studied the natural PG methods for regularized RL. However, \cite{agarwal2019theory} mainly focuses on tabular policy and log-linear, neural policy classes. \cite{cen2020fast} mainly focuses on softmax policy class.

\begin{table}
  \centering
  \vspace*{-20pt}
  \caption{ \textbf{Sample complexities} of the representative PG algorithms based on mirror descent algorithm
  for finding an $\epsilon$-stationary policy of the \textbf{nonconcave} performance function.
  Although \cite{liu2019neural,shani2020adaptive} have provided the global convergence of TRPO and PPO under
  some specific policies based on convex mirror descent, they still obtain a stationary point of nonconcave performance function.
  \textbf{Note that} our convergence analysis does not rely any specific policies. }
  \label{tab:1}   \vspace*{-10pt}
  \setlength{\tabcolsep}{5pt}
  %\resizebox{0.80\textwidth}{!}{
  \begin{tabular}{c|c|c|c}
  \hline
 \textbf{Algorithm} & \textbf{Reference} &  \textbf{Complexity}  & \textbf{Batch Size} \\ \hline
  TRPO & \citet{shani2020adaptive} & $O(\epsilon^{-4})$  & $O(\epsilon^{-2})$   \\ \hline
  Regularized TRPO & \citet{shani2020adaptive} & $O(\epsilon^{-3})$  & $O(\epsilon^{-2})$   \\ \hline
  TRPO/PPO & \citet{liu2019neural} & $O(\epsilon^{-8})$ &  $O(\epsilon^{-6})$   \\ \hline
  VRMPO & \citet{yang2019policy} & $O(\epsilon^{-3})$ &   $O(\epsilon^{-2})$  \\ \hline
  MDPO & \citet{tomar2020mirror} & Unknown  &  Unknown    \\ \hline
  BGPO & Ours &  $O(\epsilon^{-4})$  & $O(1)$  \\ \hline
  VR-BGPO & Ours & $O(\epsilon^{-3})$ & $O(1)$   \\ \hline
  %\bottomrule
  \end{tabular}
  %}
  \vspace*{-15pt}
\end{table}

Although these specific PG methods based on mirror descent iteration
have been recently studied, which are scattered in empirical and theoretical aspects respectively,
it lacks a universal framework for these PG methods without relying on
some specific RL tasks. In particular, there still does not exist the convergence analysis of
PG methods based on the mirror descent algorithm under the nonconvex setting.
Since mirror descent iteration adjusts gradient updates to fit problem geometry, and is useful in regularized RL \citep{geist2019theory},
there exists an important problem to be addressed:

\fbox{ 	\centering
\parbox{0.89\textwidth}{
\emph{ Could we design a universal policy optimization framework based on the mirror descent algorithm,
and provide its convergence guarantee under the non-convex setting ? }		
}}

In the paper, we firmly answer the above challenging question with positive solutions and
 propose an efficient Bregman gradient policy optimization framework
based on Bregman divergences and momentum techniques. In particular, we provide a convergence analysis
framework of the PG methods based on mirror descent iteration under the nonconvex setting.
In summary, our main contributions are provided as follows:
\begin{itemize}
\vspace*{-6pt}
 \item[a)] We propose an effective Bregman gradient policy optimization (BGPO) algorithm based on the basic momentum technique,
    which achieves the sample complexity of $O(\epsilon^{-4})$ for finding $\epsilon$-stationary policy
    only requiring one trajectory at each iteration.
\item[b)] We propose an accelerated Bregman gradient policy optimization (VR-BGPO) algorithm based on the variance-reduced technique of STORM  \citep{cutkosky2019momentum}.
    Moreover, we prove that the VR-BGPO reaches the best known sample complexity of $O(\epsilon^{-3})$.
\item[c)] We design a unified policy optimization framework based on mirror descent iteration and momentum techniques, and provide its convergence analysis under nonconvex setting.
    \vspace*{-6pt}
\end{itemize}
In Table \ref{tab:1} shows that sample complexities of the representative PG algorithms based on mirror descent algorithm.
\cite{shani2020adaptive,liu2019neural} have established global convergence of a mirror descent variant of PG under some pre-specified setting such as over-parameterized networks \citep{liu2019neural} by exploiting these specific problems' hidden convex nature. Without these special structures, global convergence of these methods cannot be achieved. However, our framework does not rely on any specific policy classes, and our convergence analysis only builds on the general nonconvex setting. Thus, we only prove that our methods convergence to stationary points.

\cite{geist2019theory,jin2020efficiently,lan2021policy,zhan2021policy} studied a general theory of regularized MDPs based on \textbf{policy space} such as a discrete probability space that generally is discontinuous. Since both the state and action spaces $\mathcal{S}$ and $\mathcal{A}$ generally are very large in practice, the \textbf{policy space} is large. While our methods build on \textbf{policy' parameter space} that is generally continuous Euclidean space and relatively small. Clearly, our methods and theoretical results are more practical than the results in \citep{geist2019theory,jin2020efficiently,lan2021policy,zhan2021policy}. \citep{tomar2020mirror} also proposes mirror descent PG framework based on policy parameter space, but it does not provide any theoretical results and only focuses on Bregman divergence taking form of KL divergence. While our framework can collaborate with any Bregman divergence forms.
\vspace*{-12pt}
\section{ Related Works }
\vspace*{-6pt}
In this section, we review some related works about mirror descent-based algorithms in RL and variance-reduced PG methods, respectively.
  \vspace*{-6pt}
\subsection{ Mirror Descent Algorithm in RL }
  \vspace*{-6pt}
Due to easily deal with the regularization terms, mirror descent (a.k.a., Bregman gradient) algorithm \citep{censor1992proximal,beck2003mirror} has shown significant successes in regularized RL, which is first proposed in \citep{censor1992proximal} based on Bregman distance (divergence) \citep{bregman1967relaxation,censor1981iterative}.
For example, \citet{neu2017unified} have shown both the dynamic policy programming \citep{azar2012dynamic} and TRPO \citep{schulman2015trust} algorithms
are approximate variants of mirror descent algorithm.
Subsequently, \citet{geist2019theory} have introduced a general theory of regularized MDPs based on
the convex mirror descent algorithm.
More recently, \citet{liu2019neural} have studied the global convergence properties of PPO and TRPO equipped with overparametrized neural networks
based on mirror descent iterations.
At the same time, \citet{shani2020adaptive} have analyzed the global convergence properties of TRPO in tabular policy based on
the convex mirror descent algorithm.
\citet{wang2019divergence} have proposed divergence augmented policy optimization for off-policy learning based on mirror descent algorithm.
MDPO \citep{tomar2020mirror} iteratively updates the policy
beyond the tabular RL by approximately solving a trust region problem based on
convex mirror descent algorithm.
  \vspace*{-6pt}
\subsection{ (Variance-Reduced) PG Methods }
  \vspace*{-6pt}
PG methods have been widely studied due to their stability and incremental nature in policy optimization.
For example, the global convergence properties of vanilla policy gradient method
in infinite-horizon MDPs have been recently studied in \citep{zhang2019global}.
Subsequently, \citet{zhang2020sample} have studied asymptotically global convergence properties of
the REINFORCE \citep{williams1992simple}, whose policy gradient is approximated by using a
single trajectory or a fixed size mini-batch of trajectories under soft-max parametrization
and log-barrier regularization.
To accelerate these vanilla PG methods, some faster variance-reduced PG methods
have been proposed based on the
variance-reduction techniques of SVRG \citep{johnson2013accelerating}, SPIDER \citep{fang2018spider}
and STORM \citep{cutkosky2019momentum} in stochastic  optimization.
For example, fast SVRPG \citep{papini2018stochastic,xu2019improved} algorithm have been proposed based on SVRG.
Fast HAPG \citep{shen2019hessian} and SRVR-PG \citep{xu2019improved} algorithms have been presented by using SPIDER technique.
Subsequently, the momentum-based PG methods, i.e., ProxHSPGA  \citep{pham2020hybrid} and IS-MBPG \citep{huang2020momentum}, have been developed based on variance-reduced technique of STORM/Hybrid-SGD \citep{cutkosky2019momentum,tran2019hybrid}. More recently,  \citep{ding2021global} studied the global convergence of momentum-based policy gradient methods. \citep{zhang2021convergence} proposed a truncated stochastic incremental variance-reduced policy
gradient (TSIVR-PG) method to relieve the uncheckable importance weight assumption in above variance-reduced PG methods and provided the global convergence of the TSIVR-PG under  overparameterizaiton of policy assumption.
  \vspace*{-6pt}
\section{ Preliminaries }
\vspace*{-6pt}
In the section, we will review some preliminaries of Markov decision process and policy gradients.
  \vspace*{-6pt}
\subsection{Notations}
  \vspace*{-6pt}
Let $[n]=\{1,2,\cdots,n\}$ for all $n\in \mathbb{N}_+$. For a vector $x \in \mathbb{R}^d$, let $\|x\|$ denote the $\ell_2$ norm of $x$, and  $\|x\|_p = \big(\sum_{i=1}^d|x_i|^p\big)^{1/p} \ (p\geq 1)$ denotes the $p$-norm of $x$.
For two sequences $\{a_k\}$ and $\{b_k\}$, we denote $a_k=O(b_k)$ if $a_k\leq Cb_k$ for some constant $C>0$. $\mathbb{E}[X]$ and $\mathbb{V}[X]$
denote the expectation and variance of  random variable $X$, respectively.
  \vspace*{-6pt}
\subsection{ Markov Decision Process }
  \vspace*{-6pt}
Reinforcement learning generally involves a discrete time discounted Markov Decision Process (MDP)
defined by a tuple $\{\mathcal{S},\mathcal{A},\mathbb{P},r,\gamma,\rho_0\}$.
$\mathcal{S}$ and $\mathcal{A}$ denote the state and action spaces of the agent, respectively.
$\mathbb{P}(s'|s,a) : \mathcal{S}\times \mathcal{A} \rightarrow \triangle(\mathcal{S})$ is the Markov kernel
that determines the transition probability from  the state $s$ to $s'$
under taking an action $a\in \mathcal{A}$.
$r(s,a): \mathcal{S} \times \mathcal{A}\rightarrow [-R,R] \ (R>0)$ is the reward function of $s$ and $a$,
and $\rho_0=p(s_0)$ denotes the initial state distribution.
$\gamma \in (0,1)$ is the discount factor.
Let $\pi: \mathcal{S} \rightarrow \triangle(\mathcal{A})$ be a stationary policy,
where $\triangle(\mathcal{A})$ is the set of probability distributions on $\mathcal{A}$.

Given the current state $s_t \in \mathcal{S}$, the agent executes an action $a_t \in \mathcal{A}$
following a conditional probability distribution $\pi(a_t|s_t)$, and then
the agent obtains a reward $r_t = r(s_t,a_t)$. At each time $t$, we can define the
state-action value function $Q^{\pi}(s_t,a_t)$ and state value function $V^{\pi}(s_t)$ as follows:
\begin{align}
 Q^{\pi}(s_t,a_t) = \mathbb{E}_{s_{t+1},a_{t+1},\ldots}\big[\sum_{l=0}^\infty \gamma^lr_{t+l}\big], \
 V^{\pi}(s_t) = \mathbb{E}_{a_t,s_{t+1},\ldots}\big[\sum_{l=0}^\infty \gamma^lr_{t+l}\big].
\end{align}
We also define the advantage function $A^{\pi}(s_t,a_t)=Q^{\pi}(s_t,a_t)-V^{\pi}(s_t)$.
The goal of the agent is to
find the optimal policy by maximizing the expected discounted reward
\begin{align} \label{eq:1}
 \max_{\pi} J(\pi) := \mathbb{E}_{s_0\sim \rho_0}[ V^{\pi}(s_0)].
\end{align}
Given a time horizon $H$, the agent collects a trajectory $\tau=\{s_t,a_t\}_{t=0}^{H-1}$ under any stationary policy.
Then the agent obtains a cumulative discounted reward $r(\tau) = \sum^{H-1}_{t=0}\gamma^tr(s_t,a_t)$.
Since the state and action spaces $\mathcal{S}$ and $\mathcal{A}$ are generally very large, directly solving the problem (\ref{eq:1}) is difficult.
Thus, we let the policy $\pi$ be parametrized as $\pi_{\theta}$ for the parameter $\theta \in \Theta \subseteq \mathbb{R}^d$.
Given the initial distribution $\rho_0=p(s_0)$,
the probability distribution over trajectory $\tau$ can be obtained
\begin{align}
 p(\tau|\theta) = p(s_0)\prod_{t=0}^{H-1}\mathbb{P}(s_{t+1}|s_t,a_t)\pi_{\theta}(a_t|s_t).
\end{align}
Thus, the problem (\ref{eq:1}) will be equivalent to maximize
the expected discounted trajectory reward:
\begin{align} \label{eq:2}
\max_{\theta \in \Theta} J(\theta):= \mathbb{E}_{\tau\sim p(\tau|\theta)} [r(\tau)].
\end{align}
In fact, the above objective function $J(\theta)$ has a truncation error of $O(\frac{\gamma^H}{1-\gamma})$ compared to the original infinite-horizon MDP.
  \vspace*{-6pt}
\subsection{ Policy Gradients }
  \vspace*{-6pt}
The policy gradient methods \citep{williams1992simple,sutton2000policy} are a class of effective policy-based methods to solve
the above RL problem (\ref{eq:2}). Specifically, the gradient of $J(\theta)$ with respect to $\theta$ is given as follows:
\begin{align} \label{eq:3}
\nabla J(\theta) = \mathbb{E}_{\tau\sim p(\tau|\theta)} \big[\nabla\log\big( p(\tau|\theta)\big)r(\tau)\big].
\end{align}
Given a mini-batch trajectories $\mathcal{B}=\{\tau_i\}_{i=1}^{n}$ sampled from the distribution $p(\tau |\theta)$,
the standard stochastic policy gradient ascent update at $(k+1)$-th step, defined as
\begin{align}
 \theta_{k+1} = \theta_k + \eta\nabla J_{\mathcal{B}}(\theta_k),
\end{align}
where $\eta>0$ is learning rate, and $\nabla J_{\mathcal{B}}(\theta_k) = \frac{1}{n}\sum_{i=1}^n g(\tau_i|\theta_k)$
is stochastic policy gradient.
Given $H=O(\frac{1}{1-\gamma})$ as in \citep{zhang2019global,shani2020adaptive}, $g(\tau|\theta)$ is the unbiased stochastic policy gradient of $ J(\theta)$,
\emph{i.e.}, $\mathbb{E}[g(\tau|\theta)]=\nabla J(\theta)$,
where
\begin{align}  \label{eq:3}
 g(\tau|\theta) = \big(\sum_{t=0}^{H-1} \nabla_{\theta}\log \pi_{\theta}(a_t,s_t)\big) \big(\sum_{t=0}^{H-1}\gamma^t r(s_t,a_t) \big).
\end{align}
Based on the gradient estimator in (\ref{eq:3}), we can obtain the existing well-known policy gradient estimators
such as REINFORCE \citep{williams1992simple}, policy gradient
theorem (PGT \citep{sutton2000policy}). Specifically, the REINFORCE
obtains a policy gradient estimator by adding a baseline $b$, defined as
\begin{align}
 g(\tau|\theta) = \big(\sum_{t=0}^{H-1} \nabla_{\theta}\log \pi_{\theta}(a_t,s_t)\big)  \big(\sum_{t=0}^{H-1}\gamma^t r(s_t,a_t)-b_t \big). \nonumber
\end{align}
The PGT is a version of the REINFORCE, defined as
\begin{align}
 g(\tau|\theta) = \sum_{t=0}^{H-1} \sum_{j=t}^{H-1}\big(\gamma^j r(s_j,a_j)-b_j \big)\nabla_{\theta}\log \pi_{\theta}(a_t,s_t). \nonumber
\end{align}

% Alternatively, based on the policy gradient Theorem \citep{sutton2000policy,zhang2019global}, we also have the following policy gradient form
% \begin{align} \label{eq:8}
%  \nabla J(\theta) =\mathbb{E}_{(s,a)\sim P_{\theta}(s,a)}\big[\nabla \log\pi_{\theta}(s,a)Q^{\pi_{\theta}}(s,a)\big],
% \end{align}
% where $P_{\theta}(s,a)=P_{\theta}(s)\pi_{\theta}(a|s)$ and $P_{\theta}(s)=\sum_{t=0}^{H-1}\gamma^t P(s_t=s|s_0,\pi_{\theta})$
% denotes a valid probability measure over the state $\mathcal{S}$. Here $P(s_t=s|s_0,\pi_{\theta})$
% is the probability that state $s_t=s$ given initial state $s_0$
% and policy parameter $\theta$.
% Since any function $b: \mathcal{S} \rightarrow \mathbb{R}$ is independent of action $a$,
% the policy gradient (\ref{eq:8}) is equivalent to
% \begin{align}
%  \nabla J(\theta) = \mathbb{E}_{(s,a)\sim P_{\theta}(s,a)}\big[\nabla \log\pi_{\theta}(s,a)\big(Q^{\pi_{\theta}}(s,a)\!-\!b(s)\big)\big], \nonumber
% \end{align}
% where $b(s)$ is a baseline function. When choose the state-value function $V^{\pi_{\theta}}(s)$ as a baseline function $b(s)$,
% we can obtain the advantage-based policy gradient
% \begin{align}
%  \nabla J(\theta) = \mathbb{E}_{(s,a)\sim P_{\theta}(s,a)}\big[\nabla \log\pi_{\theta}(s,a)A^{\pi_{\theta}}(s,a)\big].
% \end{align}
  \vspace*{-6pt}
\section{ Bregman Gradient Policy Optimization }
  \vspace*{-6pt}
In this section, we propose a novel Bregman gradient policy optimization framework
based on Bregman divergences and momentum techniques.
We first let $f(\theta) = -J(\theta)$, the goal of policy-based RL is to solve the problem:
$\max_{\theta\in \Theta} J(\theta) \  \Longleftrightarrow \ \min_{\theta\in \Theta} f(\theta)$,
so we have $\nabla f(\theta) = -\nabla J(\theta)$.

Assume $\psi(x)$ is a continuously-differentiable and $\nu$-strongly convex function, i.e.,
$\langle x-y, \nabla \psi(x) - \nabla \psi(y) \rangle \geq \nu \|x-y\|, \ \nu > 0$, we define a Bregman distance:
 \begin{align}
 D_{\psi}(y,x) = \psi(y)-\psi (x)-\langle \nabla \psi (x), y-x \rangle, \ \forall  x, y \in \mathbb{R}^d
\end{align}
Then given a function $h(x)$
defined on a closed convex set $\mathcal{X}$,
we define a proximal operator (a.k.a., mirror descent):
\begin{align}
 \mathcal{P}^{\psi}_{\lambda,h}(x) = \arg\min_{y\in \mathcal{X}} \big\{ h(y) + \frac{1}{\lambda}D_{\psi}(y,x) \big\},
\end{align}
where $\lambda >0$.
Based on this proximal operator  $\mathcal{P}^{\psi}_{\lambda,h}$ as in \citep{ghadimi2016mini,zhang2018convergence},
we can define a Bregman gradient of function $h(x)$ as follows:
\begin{align} \label{eq:12}
 \mathcal{B}^{\psi}_{\lambda,h}(x) = \frac{1}{\lambda}\big( x - \mathcal{P}^{\psi}_{\lambda,h}(x)\big).
\end{align}
If $\psi(x)=\frac{1}{2}\|x\|^2$ and $\mathcal{X}=\mathbb{R}^d$, $x^*$ is a stationary point of $h(x)$ if and only if $\mathcal{B}^{\psi}_{\lambda,h}(x^*) = \nabla h(x^*)=0$. Thus, this Bregman gradient can be regarded as a generalized gradient.
  \vspace*{-6pt}
\subsection{ BGPO Algorithm }
  \vspace*{-6pt}
In the subsection, we propose a Bregman gradient policy optimization (BGPO) algorithm based on the basic momentum technique.
The pseudo code of BGPO Algorithm is provided in Algorithm \ref{alg:1}.

\begin{algorithm}[tb]
\caption{ BGPO Algorithm }
\label{alg:1}
\begin{algorithmic}[1] %[1] enables line numbers
\STATE {\bfseries Input:}  Total iteration $K$, tuning parameters $\{\lambda,b,m,c\}$ and mirror mappings $\big\{\psi_k\big\}_{k=1}^K$
are $\nu$-strongly convex functions; \\
\STATE {\bfseries Initialize:} $\theta_1 \in \Theta$, and sample a trajectory $\tau_1$ from $p(\tau |\theta_1)$,
and compute $u_1 = -g(\tau_1|\theta_1)$;\\
\FOR{$k = 1, 2, \ldots, K$}
\STATE Update  $\tilde{\theta}_{k+1} = \arg\min_{\theta\in \Theta} \big\{ \langle u_k, \theta\rangle
+ \frac{1}{\lambda} D_{\psi_k}(\theta,\theta_k) \big\}$; \\
\STATE Update  $\theta_{k+1} = \theta_k + \eta_k(\tilde{\theta}_{k+1}-\theta_k)$ with $\eta_k = \frac{b}{(m+k)^{1/2}}$;
\STATE Sample a trajectory $\tau_{k+1}$ from $p(\tau |\theta_{k+1})$, and compute
$u_{k+1} = -\beta_{k+1} g(\tau_{k+1}|\theta_{k+1}) + (1-\beta_{k+1})u_k$ with $\beta_{k+1} = c\eta_k$; \\
\ENDFOR
\STATE {\bfseries Output:}  $\theta_{\zeta}$ chosen uniformly random from $\{\theta_k\}_{k=1}^{K}$.
\end{algorithmic}
\end{algorithm}

In Algorithm \ref{alg:1}, the step 4 uses the stochastic Bregman gradient descent (a.k.a., stochastic mirror descent)
to update the parameter $\theta$. Let $h(\theta)=\langle\theta, u_k\rangle$ be the first-order approximation of
function $f(\theta)$ at $\theta_k$, where $u_k$ is an  approximated gradient of function $f(\theta)$ at $\theta_k$.
By the step 4 of Algorithm \ref{alg:1} and the above equality (\ref{eq:12}), we have
\begin{align}
 \mathcal{B}^{\psi_k}_{\lambda,h}(\theta_k) = \frac{1}{\lambda}\big(\theta_k - \tilde{\theta}_{k+1}\big),
\end{align}
where $\lambda>0$.
Then by the step 5 of  Algorithm \ref{alg:1}, we have
\begin{align} \label{eq:14}
 \theta_{k+1} = \theta_k - \lambda\eta_k\mathcal{B}^{\psi_k}_{\lambda,h}(\theta_k),
\end{align}
where $0<\eta_k\leq 1$. Due to the convexity of set $\Theta \subseteq \mathbb{R}^d$ and
$\theta_1\in \Theta$, we choose the parameter $\eta_k\in (0,1]$ to ensure the updated sequence
$\{\theta_k\}_{k=1}^K$ in $\Theta$.

In fact, our BGPO algorithm unifies many popular policy optimization algorithms.
When the mirror mappings $\psi_k(\theta)=\frac{1}{2}\|\theta\|^2$ for $\forall k\geq 1$,
the update (\ref{eq:14}) will be equivalent to
a classic policy gradient iteration. Then our BGPO algorithm will become a momentum version of the
policy gradient algorithms  \citep{sutton2000policy,zhang2019global}.
Given $\psi_k(\theta)=\frac{1}{2}\|\theta\|^2$ and $\beta_k=1$, i.e.,  $u_k=-g(\tau_k|\theta_k)$,
we have $\mathcal{B}^{\psi_k}_{\lambda,h}(\theta_k)=-g(\tau_k|\theta_k)$ and
\begin{align}
 \theta_{k+1} = \theta_k + \lambda\eta_kg(\tau_k|\theta_k).
\end{align}
When the mirror mappings $\psi_k(\theta)=\frac{1}{2}\theta^TF(\theta_k)\theta$ with
$F(\theta_k)=\mathbb{E}\big[\nabla_{\theta}\pi_{\theta_k}(s,a)\big(\nabla_{\theta}\pi_{\theta_k}(s,a)\big)^T\big]$, the update (\ref{eq:14}) will be equivalent to
a natural policy gradient iteration. Then our BGPO will become a momentum version of  natural policy gradient algorithms \citep{kakade2001natural,liu2020improved}.
Given $\psi_k(\theta)=\frac{1}{2}\theta^TF(\theta_k)\theta$, $\beta_k=1$, i.e.,  $u_k=-g(\tau_k|\theta_k)$,
we have $\mathcal{B}^{\psi_k}_{\lambda,h}(\theta_k)=-F(\theta_k)^+g(\tau_k|\theta_k)$ and
\begin{align}
 \theta_{k+1} = \theta_k + \lambda\eta_kF(\theta_k)^+g(\tau_k|\theta_k),
\end{align}
where $F(\theta_k)^+$ denotes the Moore-Penrose pseudoinverse of the Fisher information matrix $F(\theta_k)$.
When given the mirror mapping $\psi_k(\theta) = \sum_{s\in \mathcal{S}} \pi_{\theta}(s)\log(\pi_{\theta}(s))$, i.e., Boltzmann-Shannon entropy function \citep{shannon1948mathematical} and $\Theta = \big\{ \theta \in \mathbb{R}^d \ | \ \sum_{s\in \mathcal{S}}\pi_{\theta}(s)=1 \big\}$, we have
$D_{\psi_k}(\theta,\theta_k) = \mbox{KL}\big(\pi_{\theta}(s),\pi_{\theta_k}(s)\big) = \sum_{s\in \mathcal{S}}\pi_{\theta}(s)\log\big(\frac{\pi_{\theta}(s)}{\pi_{\theta_k}(s)}\big) $, which is the KL divergence.
Then our BGPO will become a momentum version of mirror descent policy optimization \citep{tomar2020mirror}.
  \vspace*{-6pt}
\subsection{ VR-BGPO Algorithm }
  \vspace*{-6pt}
In the subsection, we propose a faster variance-reduced Bregman gradient policy optimization (VR-BGPO) algorithm
based on a variance-reduced technique.
The pseudo code of VR-BGPO algorithm is provided in Algorithm \ref{alg:2}.

Consider the problem (\ref{eq:2}) is non-oblivious that the distribution $p(\tau|\theta)$
depends on the variable $\theta$ varying through the whole optimization procedure,
we apply the importance sampling weight \citep{papini2018stochastic,xu2019improved} in estimating our policy gradient $u_{k+1}$, defined as
\begin{align} \label{eq:5}
  w(\tau_{k+1}|\theta_k,\theta_{k+1})
  = \frac{p(\tau_{k+1}|\theta_{k})}{p(\tau_{k+1}|\theta_{k+1})} = \prod_{t=0}^{H-1}\frac{\pi_{\theta_{k}}(a_t|s_t)}{\pi_{\theta_{k+1}}(a_t|s_t)}. \nonumber
\end{align}
Except for different stochastic policy gradients $\{u_k\}$ and tuning parameters $\{\eta_k,\beta_k\}$ using in Algorithms \ref{alg:1} and
\ref{alg:2}, the steps 4 and 5 in these algorithms for updating parameter $\theta$ are the same. Interestingly, when choosing mirror mapping $\psi_k(\theta)=\frac{1}{2}\|\theta\|^2$, our VR-BGPO algorithm will reduce to a non-adaptive version of IS-MBPG algorithm \citep{huang2020momentum}.

\begin{algorithm}[tb]
\caption{ VR-BGPO Algorithm}
\label{alg:2}
\begin{algorithmic}[1] %[1] enables line numbers
\STATE {\bfseries Input:}  Total iteration $K$, tuning parameters $\{\lambda,b,m,c\}$ and mirror mappings $\big\{\psi_k\big\}_{k=1}^K$
are $\nu$-strongly convex functions; \\
\STATE {\bfseries Initialize:}  $\theta_1 \in \Theta$, and sample a trajectory $\tau_1$ from $p(\tau |\theta_1)$, and compute $u_1 =  -g(\tau_1|\theta_1)$;\\
\FOR{$k = 1, 2, \ldots, K$}
\STATE Update  $\tilde{\theta}_{k+1} = \arg\min_{\theta\in \Theta} \big\{ \langle u_k, \theta\rangle + \frac{1}{\lambda} D_{\psi_k}(\theta,\theta_k) \big\}$; \\
\STATE Update $\theta_{k+1} = \theta_k + \eta_k(\tilde{\theta}_{k+1}-\theta_k)$ with $\eta_k = \frac{b}{(m+k)^{1/3}}$;
\STATE Sample a trajectory $\tau_{k+1}$ from $p(\tau |\theta_{k+1})$, and compute $u_{k+1} = -\beta_{k+1} g(\tau_{k+1}|\theta_{k+1}) + (1-\beta_{k+1})\big[u_k - g(\tau_{k+1} | \theta_{k+1})+ w(\tau_{k+1}|\theta_k,\theta_{k+1}) g(\tau_{k+1}|\theta_k)\big]$
with $\beta_{k+1} = c\eta_k^2$; \\
\ENDFOR
\STATE {\bfseries Output:}  $\theta_{\zeta}$ chosen uniformly random from $\{\theta_k\}_{k=1}^{K}$.
\end{algorithmic}
\end{algorithm}
  \vspace*{-6pt}
\section{Convergence Analysis}
  \vspace*{-6pt}
In this section, we will analyze the convergence properties of our BGPO and VR-BGPO algorithms.
All related proofs are provided in the Appendix \ref{Appendix:A}. Here we use the standard convergence metric $\|\mathcal{B}^{\psi_k}_{\lambda,\langle\theta,\nabla f(\theta_k)\rangle}(\theta)\|$
used in \citep{zhang2018convergence,yang2019policy} to evaluate the convergence Bregman gradient-based (a.k.a., mirror descent) algorithms.
To give the convergence analysis, we first give some standard assumptions.
\begin{assumption}
For function $\log\pi_{\theta}(a|s)$, its gradient and Hessian matrix are bounded, \emph{i.e.,} there exist constants $C_g, C_h >0$ such that
$ \|\nabla_{\theta}\log\pi_{\theta}(a|s)\| \leq C_g, \ \|\nabla^2_{\theta}\log\pi_{\theta}(a|s)\| \leq C_h$.
\end{assumption}
\begin{assumption}
Variance of stochastic gradient $g(\tau|\theta)$ is bounded, \emph{i.e.,} there exists a constant $\sigma >0$, for all $\pi_{\theta}$
such that $\mathbb{V}(g(\tau|\theta)) = \mathbb{E}\|g(\tau|\theta)-\nabla J(\theta)\|^2 \leq \sigma^2$.
\end{assumption}
\begin{assumption}
For importance sampling weight $w(\tau|\theta_1,\theta_2)=p(\tau|\theta_1)/p(\tau|\theta_2)$, its variance is bounded, \emph{i.e.,} there exists a constant $W >0$, it follows $\mathbb{V}(w(\tau|\theta_1,\theta_2)) \leq W$ for any $\theta_1, \theta_2 \in \mathbb{R}^d$ and $\tau\sim p(\tau|\theta_2)$.
\end{assumption}
\begin{assumption}
The function $J(\theta)$ has an upper bound in $\Theta$, \emph{i.e.,} $J^* = \sup_{\theta \in \Theta} J(\theta) < +\infty$.
\end{assumption}
Assumptions 1 and 2 are commonly used in
the PG algorithms \citep{papini2018stochastic,xu2019improved,xu2019sample}.
Assumption 3 is widely used in the study of variance reduced PG algorithms
\citep{papini2018stochastic,xu2019improved}.
In fact, the bounded importance sampling weight might be
violated in some cases such as using neural networks as the policy.
Thus, we can clip this importance sampling weights to guarantee the effectiveness of our algorithms
as in \citep{papini2018stochastic}.
At the same time, the importance weights actually also have some nice properties, e.g., in soft-max policy it is bounded by $e^{c\|\theta_1-\theta_2\|^2}$ for all $\theta_1,\theta_2\in \Theta$.
More recently, \citep{zhang2021convergence} used a simple truncated update to relieve this uncheckable importance weight assumption.
Assumption 4 guarantees the feasibility of the problem (\ref{eq:2}).  Note that Assumptions 2 and 4 are satisfied automatically given Assumption 1 and the fact that all the rewards are bounded, i.e., $|r(s,a)|\leq R$ for any $s\in \mathcal{S}$ and $a\in \mathcal{A}$. For example, due to $|r(s,a)|\leq R$, we have $|J(\theta)| \leq \frac{R}{1-\gamma}$. So we have $J^* = \frac{R}{1-\gamma}$.
  \vspace*{-6pt}
\subsection{Convergence Analysis of BGPO Algorithm}
  \vspace*{-6pt}
In the subsection, we provide convergence properties of the BGPO algorithm.
The detailed proof is provided in Appendix \ref{Appendix:A1}.
\begin{theorem} \label{th:1}
 Assume the sequence $\{\theta_k\}_{k=1}^K$ be generated from Algorithm \ref{alg:1}.
 Let $\eta_k=\frac{b}{(m+k)^{1/2}}$ for all $k\geq 1$, $0< \lambda \leq \frac{\nu m^{1/2}}{9Lb}$, $b>0$, $\frac{8L\lambda}{\nu} \leq c \leq \frac{m^{1/2}}{b}$,
 and $m \geq \max\{b^2,(cb)^2\}$, we have
 \begin{align}
  \frac{1}{K} \sum_{k=1}^K \mathbb{E} \|\mathcal{B}^{\psi_k}_{\lambda,\langle \nabla f(\theta_k), \theta\rangle}(\theta_k)\|
  \leq \frac{2\sqrt{2M}m^{1/4}}{K^{1/2}} + \frac{2\sqrt{2M}}{K^{1/4}}, \nonumber
 \end{align}
 where $M = \frac{J^* - J(\theta_1)}{\nu\lambda b} + \frac{\sigma^2}{\nu\lambda Lb} + \frac{m\sigma^2}{\nu \lambda Lb}\ln(m+K)$.
 \end{theorem}
\begin{remark}
Without loss of generality, let $b=O(1)$, $m=O(1)$ and $\lambda=O(1)$, we have $M=O(\ln(m+K))=\tilde{O}(1)$.
Theorem \ref{th:1} shows that the BGPO algorithm has a convergence rate
of $\tilde{O}(\frac{1}{K^{1/4}})$.
Let $K^{-\frac{1}{4}} \leq \epsilon$, we have $K=\tilde{O}(\epsilon^{-4})$.
Since the BGPO algorithm only needs one trajectory
to estimate the stochastic policy gradient at each iteration and runs $K$ iterations,
it has the sample complexity of $1\cdot K = O(\epsilon^{-4})$
for finding an $\epsilon$-stationary point.
\end{remark}
  \vspace*{-6pt}
\subsection{Convergence Analysis of VR-BGPO Algorithm}
  \vspace*{-6pt}
In the subsection, we give convergence properties of the VR-BGPO algorithm.
The detailed proof is provided in Appendix \ref{Appendix:A2}.
\begin{theorem} \label{th:2}
Suppose the sequence $\{\theta_k\}_{k=1}^K$ be generated from Algorithm \ref{alg:2}. Let $\eta_k = \frac{b}{(m+k)^{1/3}}$
 for all $k \geq 0$, $0< \lambda \leq \frac{\nu m^{1/3}}{5\hat{L}b}$, $b>0$,
 $c \in \big[\frac{2}{3b^3} + \frac{20\hat{L}^2 \lambda^2}{\nu^2}, \frac{m^{2/3}}{b^2}\big]$ and $m\geq \max\big(2,b^3,(cb)^3,(\frac{5}{6b})^{2/3}\big)$,
 we have
  \begin{align}
  \frac{1}{K} \sum_{k=1}^K \mathbb{E} \|\mathcal{B}^{\psi_k}_{\lambda,\langle \nabla f(\theta_k), \theta\rangle}(\theta_k)\|
  \leq \frac{2\sqrt{2M'}m^{1/6}}{K^{1/2}} + \frac{2\sqrt{2M'}}{K^{1/3}},
 \end{align}
 where $M'=\frac{J^*-J(\theta_1)}{b\nu\lambda} + \frac{m^{1/3}\sigma^2}{16b^2\hat{L}^2\lambda^2}
   + \frac{c^2\sigma^2 b^2}{8\hat{L}^2\lambda^2}$, $\hat{L}^2 = L^2 + 2G^2C^2_w$, $G=C_gR/(1-\gamma)^2$ and $C_w = \sqrt{H(2HC_g^2+C_h)(W+1)}$.
\end{theorem}
\begin{remark}
Without loss of generality, let $b=O(1)$, $m=O(1)$ and $\lambda=O(1)$, we have $M=O(\ln(m+K))=\tilde{O}(1)$.
Theorem \ref{th:2} shows that the VR-BGPO algorithm has a convergence rate of $\tilde{O}(\frac{1}{K^{1/3}})$.
Let $K^{-\frac{1}{3}} \leq \epsilon$, we have $K=(\epsilon^{-3})$.
Since the VR-BGPO algorithm only needs one trajectory
to estimate the stochastic policy gradient at each iteration and runs $K$ iterations,
it reaches a lower sample complexity of $1\cdot K = \tilde{O}(\epsilon^{-3})$
for finding an $\epsilon$-stationary point.
\end{remark}
  \vspace*{-6pt}
\section{Experiments}
  \vspace*{-6pt}
In this section, we conduct some RL tasks to verify the effectiveness of our methods. We first study the effect of different choices of Bregman divergences with our algorithms (BGPO and VR-BGPO), and then we compare our VR-BGPO algorithm with other state-of-the-art methods such as TRPO~ \citep{schulman2015trust}, PPO~\citep{schulman2017proximal}, ProxHSPGA~\citep{pham2020hybrid}, VRMPO~\citep{yang2019policy}, and MDPO~\citep{tomar2020mirror}. Our code is available at \url{https://github.com/gaosh/BGPO}.
  \vspace*{-6pt}
\subsection{Effects of Bregman Divergences}\label{effects_BD}
  \vspace*{-6pt}
\begin{figure*}[t]
	\centering
  \vspace*{-15pt}
	\subfloat[CartPole-v1]{
			\includegraphics[width=.25\textwidth]{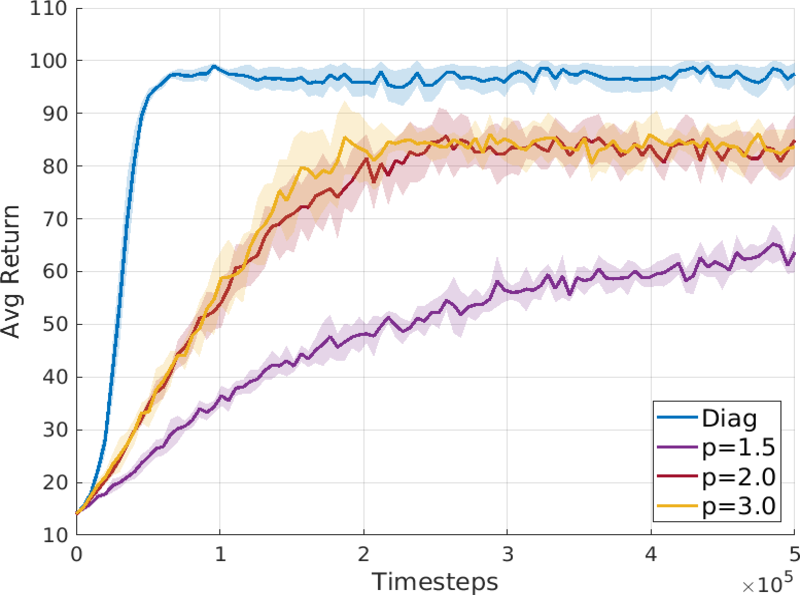}} \quad
	\subfloat[Acrobat-v1]{
			\includegraphics[width=.25\textwidth]{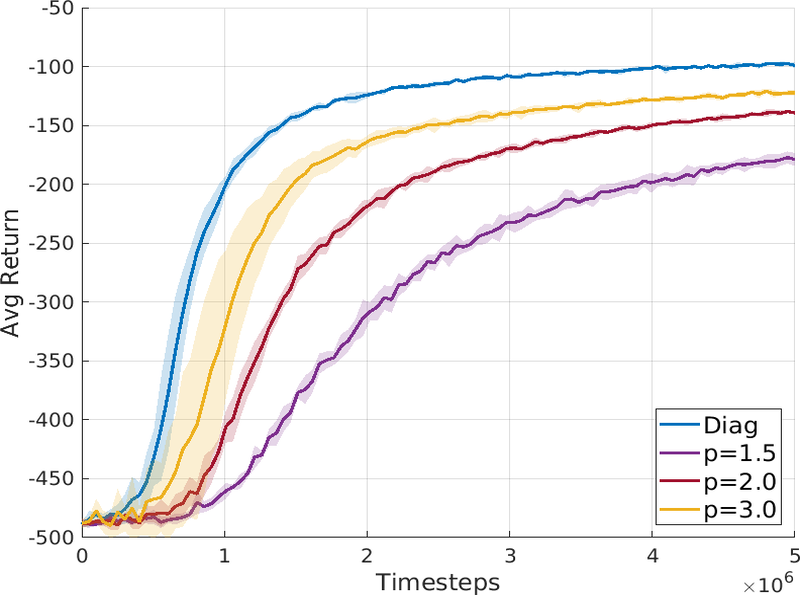}} \quad
	\subfloat[MountainCarContinuous]{
			\includegraphics[width=.25\textwidth]{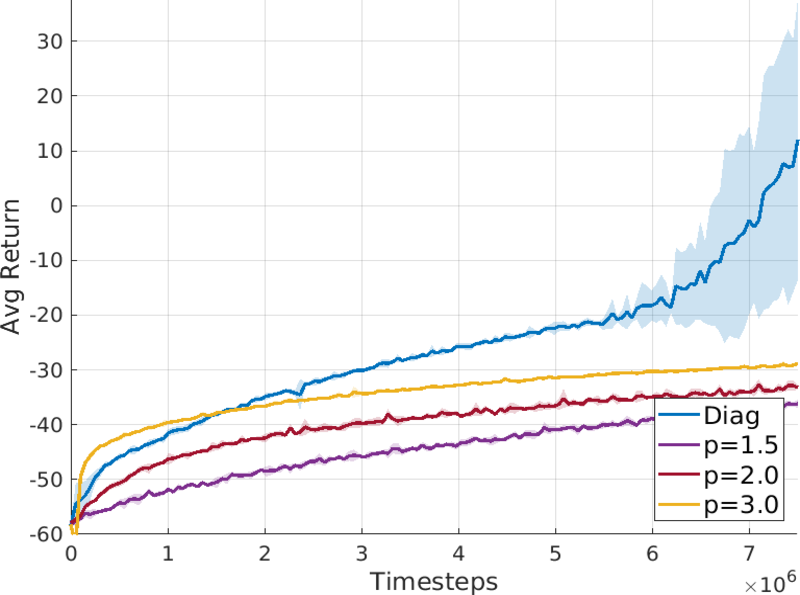}}
\vspace*{-8pt}			
	\caption{Effects of two Bregman Divergences: $l_p$-norm and diagonal term (Diag). }
	\label{fig:diff-settings}
\end{figure*}

\begin{figure*}[t]
	\centering
  \vspace*{-10pt}
	\subfloat[CartPole-v1]{
			\includegraphics[width=.25\textwidth]{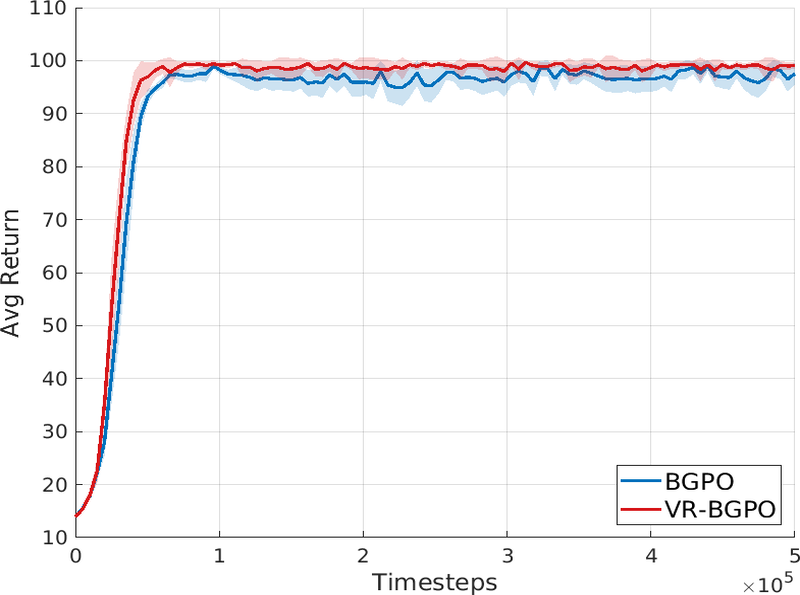}} \quad
	\subfloat[Acrobat-v1]{
			\includegraphics[width=.25\textwidth]{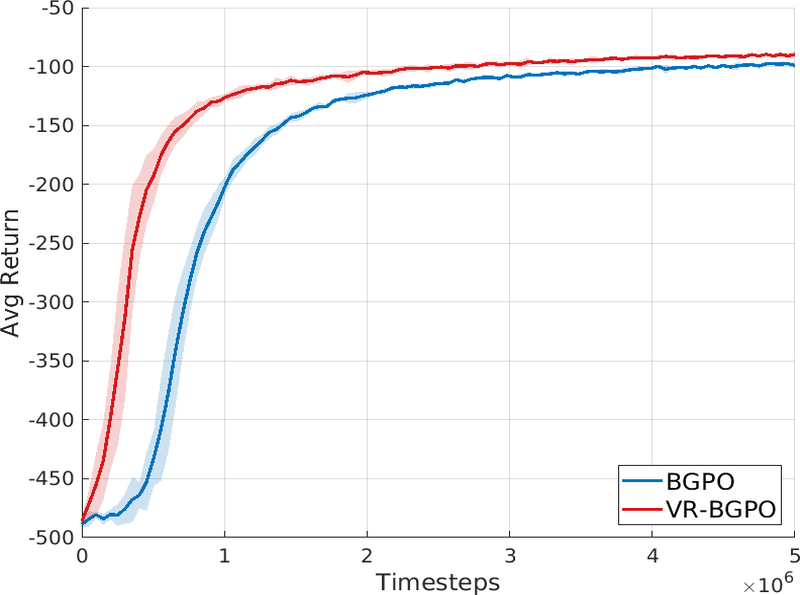}} \quad
	\subfloat[MountainCarContinuous]{
			\includegraphics[width=.25\textwidth]{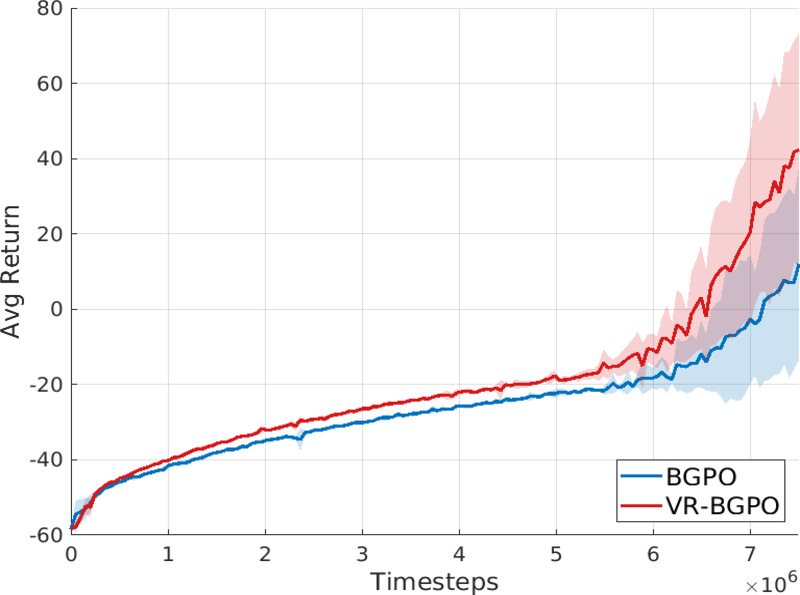}}
\vspace*{-8pt}			
	\caption{Comparison between BGPO and VR-BGPO on different environments. }
	\label{fig:vr-effect}
\end{figure*}

\begin{figure*}[t]
	\centering
  \vspace*{-6pt}
  	\subfloat[Pendulum-v2]{
			\includegraphics[width=.25\textwidth]{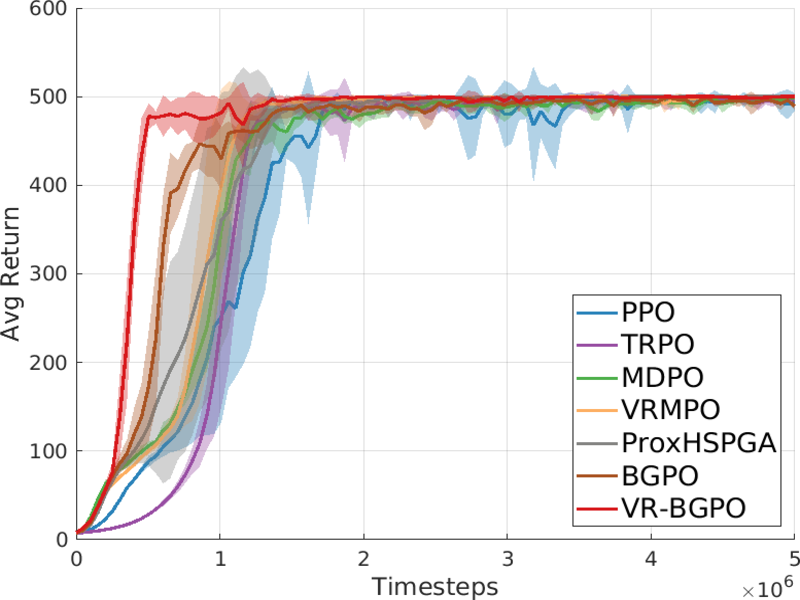}}
			\quad
	\subfloat[DoublePendulum-v2]{
			\includegraphics[width=.25\textwidth]{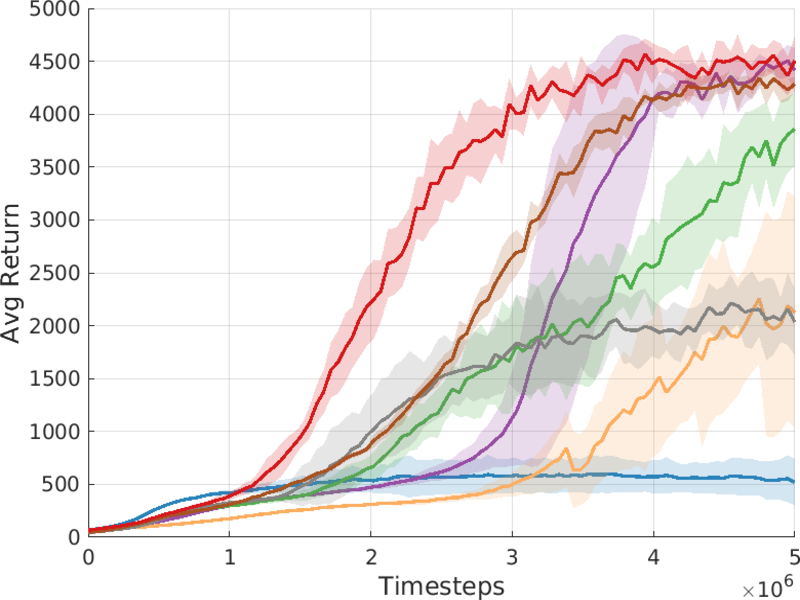}}
			\quad
	\subfloat[Walker2d-v2]{
			\includegraphics[width=.25\textwidth]{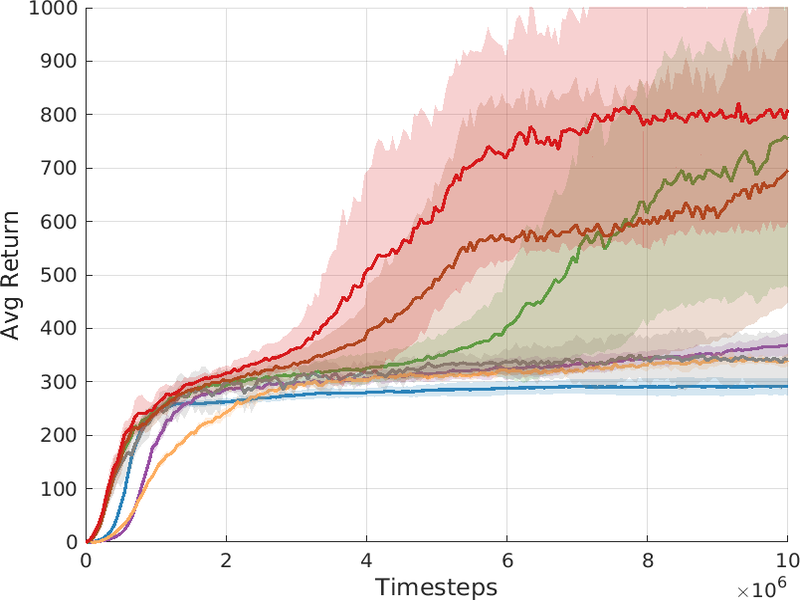}} \\
    \subfloat[Swimmer-v2]{
			\includegraphics[width=.25\textwidth]{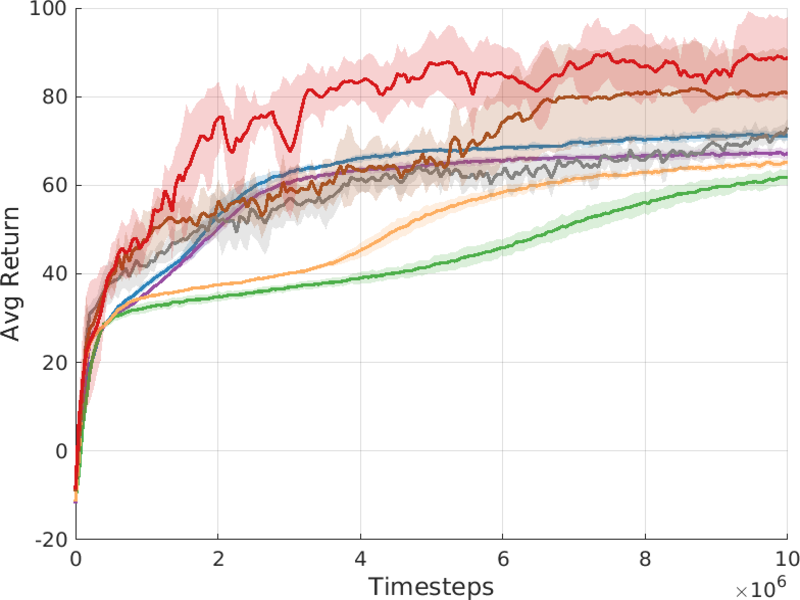} }
			\quad
	\subfloat[Reacher-v2]{
			\includegraphics[width=.25\textwidth]{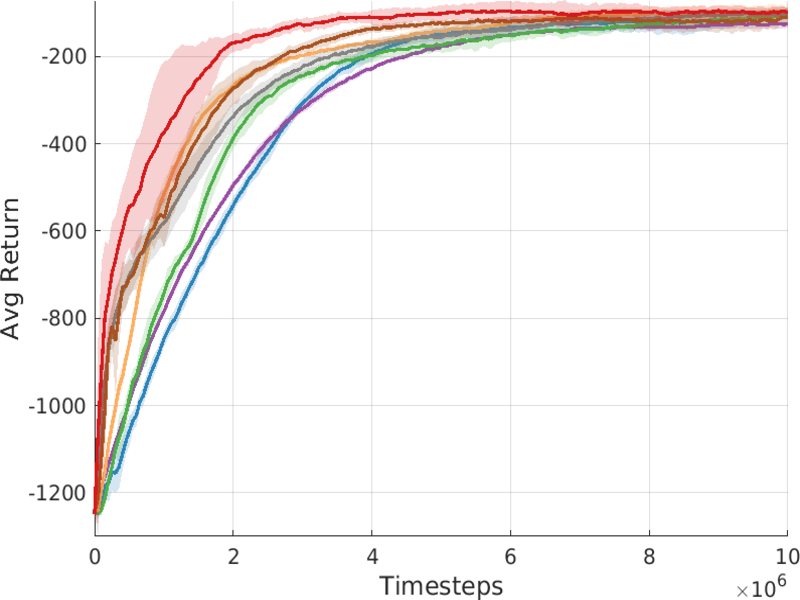}}
			\quad
	\subfloat[HalfCheetah-v2]{
			\includegraphics[width=.25\textwidth]{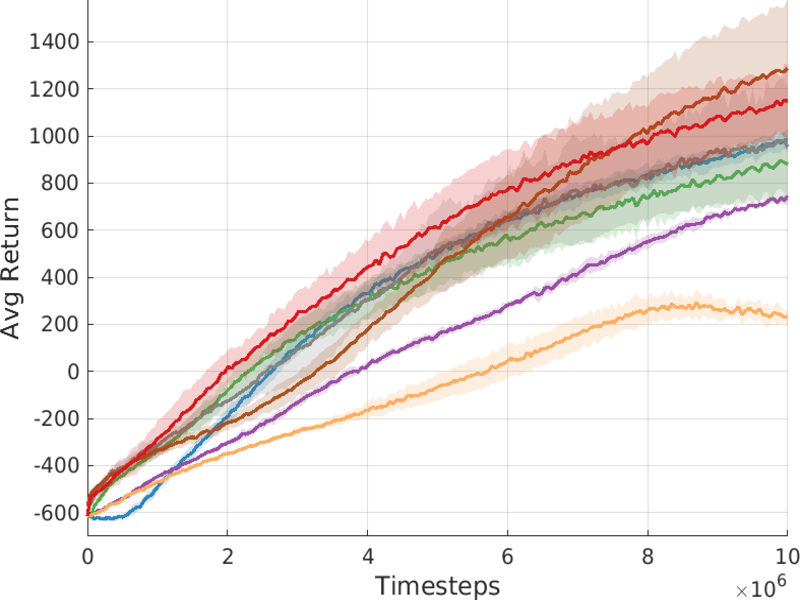}}
\vspace*{-8pt}			
	\caption{Experimental results of our algorithms and other baseline algorithms on six environments. }
	\label{fig:compare_all}
\vspace*{-15pt}
\end{figure*}

In the subsection, we examine how different Bregman divergences affect the performance of our algorithms. In the first setting, we let mirror mapping $\psi_k(x)=\|x\|_p \ (p\geq 1)$ with different $p$ to test the performance our algorithms. Let $\psi^{*}_k(y)=(\sum_{i=1}^d|y_i|^q)^{\frac{1}{q}}$ be the conjugate mapping of $\psi_k(x)$, where $p^{-1} + q^{-1}=1,\ p,q>1$. According to ~\citep{beck2003mirror}, when $\Theta=\mathbb{R}^d$, the update of $\tilde{\theta}_{k+1}$ in our algorithms can be calculated by
$\tilde{\theta}_{k+1} = \nabla\psi^{*}_k(\nabla\psi_k(\theta_k) + \lambda u_k)$,
where $\nabla\psi_k(x_j)$ and $\nabla\psi_k^{*}(y_j)$ are $p$-norm link functions, and $\nabla\psi_k(x_j) = \frac{\text{sign}(x_j){|x_j|}^{p-1}}{\|x\|^{p-2}_p}$, $\nabla\psi_k^{*}(y_j) = \frac{\text{sign}(y_j){|y_j|}^{q-1}}{\|y\|^{q-2}_q}$, and $j$ is the coordinate index of $x$ and $y$.
In the second setting, we apply diagonal term on the mirror mapping $\psi_k(x) = \frac{1}{2} x^T M_k x$, where $M_k$ is a diagonal matrix with positive values. In the experiments, we generate $H_k = \mbox{diag}(\sqrt{v_k} + \alpha )$, $v_k = \beta v_{k-1} + (1-\beta) u_k^2$, and $\alpha > 0, \beta\in (0,1)$, as in Super-Adam algorithm \citep{kingma2014adam,huang2021super}. Then we have $D_{\psi_k}(y,x) = \frac{1}{2} (y-x)^T H_k (y-x)$. Under this setting, the update of $\tilde{\theta}_{k+1}$ can also be analytically solved $\tilde{\theta}_{k+1}  = \theta_k - \lambda H_k^{-1} u_k$.

To test the effectiveness of two different Bregman divergences, we evaluate them on three classic control environments from gym~\cite{gym}: CartPole-v1, Acrobat-v1, and MountainCarContinuous-v0. In the experiment, categorical policy is used for CartPole and Acrobot environments, and Gaussian policy is used for MountainCar. Gaussian value functions are used in all settings. All policies and value functions are parameterized by multilayer perceptrons (MLPs). For a fair comparison, all settings use the same initialization for policies. We run each setting five times and plot the mean and variance of average returns.
For $l_p$-norm mapping, we test three different values of $p = (1.50, 2.0, 3.0)$. For diagonal mapping,
we set $\beta = 0.999$ and $\alpha = 10^{-8}$. We set hyperparameters $\{b, m, c\}$ to be the same.
$\lambda$ still needs to be tuned for different $p$ to achieve relatively good performance.
For simplicity, we use BGPO-Diag to represent BGPO with diagonal mapping, and we use BGPO-$l_p$ to represent BGPO with $l_p$-norm mapping.
Details about the setup of environments and hyperparameters are provided in the
Appendix \ref{appendix:B}.

From Fig.~\ref{fig:diff-settings}, we can find that BGPO-Diag largely outperforms BGPO-$l_p$ with different choices of $p$.
The parameter tuning of BGPO-$l_p$ is much more difficult than BGPO-Diag because each $p$ requires an individual $\lambda$ to achieve the desired performance.
  \vspace*{-6pt}
\subsection{Comparison between BGPO and VR-BGPO} \label{vr_effects}
  \vspace*{-6pt}
To understand the effectiveness of variance reduced technique used in our VR-BGPO algorithm, we compare BGPO and VR-BGPO using the same settings introduced in section.~\ref{effects_BD}. Both algorithms use the diagonal mapping for $\psi$, since it performs much better than $l_p$-norm.
From Fig.~\ref{fig:vr-effect} given in the
Appendix \ref{appendix:B}, we can see that VR-BGPO can outperform BGPO in all three environments.
In CartPole, both algorithms converge very fast and have similar performance, and VR-BGPO is more stable than BGPO.
The advantage of VR-BGPO becomes large in Acrobot and MountainCar environments, probably because the task is more difficult compared to CartPole.
\vspace*{-6pt}
\subsection{Compare to other Methods} \label{other_methods}
\vspace*{-6pt}
In this subsection, we apply our BGPO and VR-BGPO algorithms to compare with the other methods.
For our BGPO and VR-BGPO, we use diagonal mapping for $\psi$. For VRMPO, we follow their implementation and use $l_p$-norm for $\psi$. For MDPO, $\psi$ is the negative Shannon entropy, and the Bregman divergence becomes KL-divergence.

To evaluate the performance of these algorithms, we test them on six gym~\citep{gym} environments with continuous control tasks: Inverted-DoublePendulum-v2, Walker2d-v2, Reacher-v2, Swimmer-v2, Inverted-Pendulum-v2 and HalfCheetah-v2. We use Gaussian policies and Gaussian value functions for all environments, and both of them are parameterized by MLPs. To ensure a fair comparison, all policies use the same initialization. For TRPO and PPO, we use the implementations provided by garage~\citep{garage}. We carefully implement MDPO and VRMPO following the description provided by the original papers. All methods include our method, are implemented with garage~\citep{garage} and pytorch~\citep{paszke2019pytorch}. We run all algorithms ten times on each environment and report the mean and variance of average returns. Details about the setup of environments and hyperparameters
are also provided in the Appendix \ref{appendix:B}.

From  Fig.~\ref{fig:compare_all}, we can find that our VR-BGPO method  consistently outperforms all the other methods.
Our BGPO basically  reaches the second best performances.
From the results of our BGPO, we can find that using a proper Bregman (mirror) distance can improve performances of the PG methods.
From the results of our VR-BGPO, we can find that using a proper variance-reduced technique can further improve performances of the BGPO.
% Specifically, our method achieves the best mean average return in InvertedDoublePendulum, Swimmer, and HalfCheetah.
% Our method converges faster than other methods in all environments, especially in InvertedPendulum, InvertedDoublePendulum, and Reacher.
ProxHSPGA can reach some relatively good performances by using the variance reduced technique.
MDPO can achieve good results in some environments, but it can not outperform PPO or TRPO in Swimmer and InvertedDoublePendulum.
VRMPO only outperforms PPO and TRPO in Reacher and InvertedDoublePendulum. The undesirable performance of VRMPO is probably
because it uses $l_p$ norm for $\psi$, which requires careful tuning of learning rate.
  \vspace*{-8pt}
\section{Conclusion}
  \vspace*{-6pt}
In the paper, we proposed a novel Bregman gradient policy optimization framework for reinforcement learning based on Bregman divergences and momentum techniques. Moreover, we studied convergence properties of the proposed methods under the nonconvex setting.

% Acknowledgements should go at the end, before appendices and references
\vspace*{-8pt}
\section*{Acknowledgment}
\vspace*{-6pt}
This work was partially supported by NSF IIS 1845666, 1852606, 1838627, 1837956, 1956002, OIA 2040588.


%\vskip 0.2in

\bibliography{BMPO}

\begin{thebibliography}{52}
\providecommand{\natexlab}[1]{#1}
\providecommand{\url}[1]{\texttt{#1}}
\expandafter\ifx\csname urlstyle\endcsname\relax
  \providecommand{\doi}[1]{doi: #1}\else
  \providecommand{\doi}{doi: \begingroup \urlstyle{rm}\Url}\fi

\bibitem[Agarwal et~al.(2019)Agarwal, Kakade, Lee, and
  Mahajan]{agarwal2019theory}
Alekh Agarwal, Sham~M Kakade, Jason~D Lee, and Gaurav Mahajan.
\newblock On the theory of policy gradient methods: Optimality, approximation,
  and distribution shift.
\newblock \emph{arXiv preprint arXiv:1908.00261}, 2019.

\bibitem[Azar et~al.(2012)Azar, G{\'o}mez, and Kappen]{azar2012dynamic}
Mohammad~Gheshlaghi Azar, Vicen{\c{c}} G{\'o}mez, and Hilbert~J Kappen.
\newblock Dynamic policy programming.
\newblock \emph{The Journal of Machine Learning Research}, 13\penalty0
  (1):\penalty0 3207--3245, 2012.

\bibitem[Beck \& Teboulle(2003)Beck and Teboulle]{beck2003mirror}
Amir Beck and Marc Teboulle.
\newblock Mirror descent and nonlinear projected subgradient methods for convex
  optimization.
\newblock \emph{Operations Research Letters}, 31\penalty0 (3):\penalty0
  167--175, 2003.

\bibitem[Bregman(1967)]{bregman1967relaxation}
Lev~M Bregman.
\newblock The relaxation method of finding the common point of convex sets and
  its application to the solution of problems in convex programming.
\newblock \emph{USSR computational mathematics and mathematical physics},
  7\penalty0 (3):\penalty0 200--217, 1967.

\bibitem[Brockman et~al.(2016)Brockman, Cheung, Pettersson, Schneider,
  Schulman, Tang, and Zaremba]{gym}
Greg Brockman, Vicki Cheung, Ludwig Pettersson, Jonas Schneider, John Schulman,
  Jie Tang, and Wojciech Zaremba.
\newblock Openai gym, 2016.

\bibitem[Cen et~al.(2020)Cen, Cheng, Chen, Wei, and Chi]{cen2020fast}
Shicong Cen, Chen Cheng, Yuxin Chen, Yuting Wei, and Yuejie Chi.
\newblock Fast global convergence of natural policy gradient methods with
  entropy regularization.
\newblock \emph{arXiv preprint arXiv:2007.06558}, 2020.

\bibitem[Censor \& Lent(1981)Censor and Lent]{censor1981iterative}
Yair Censor and Arnold Lent.
\newblock An iterative row-action method for interval convex programming.
\newblock \emph{Journal of Optimization theory and Applications}, 34\penalty0
  (3):\penalty0 321--353, 1981.

\bibitem[Censor \& Zenios(1992)Censor and Zenios]{censor1992proximal}
Yair Censor and Stavros~Andrea Zenios.
\newblock Proximal minimization algorithm withd-functions.
\newblock \emph{Journal of Optimization Theory and Applications}, 73\penalty0
  (3):\penalty0 451--464, 1992.

\bibitem[Cortes et~al.(2010)Cortes, Mansour, and Mohri]{cortes2010learning}
Corinna Cortes, Yishay Mansour, and Mehryar Mohri.
\newblock Learning bounds for importance weighting.
\newblock In \emph{Advances in neural information processing systems}, pp.\
  442--450, 2010.

\bibitem[Cutkosky \& Orabona(2019)Cutkosky and Orabona]{cutkosky2019momentum}
Ashok Cutkosky and Francesco Orabona.
\newblock Momentum-based variance reduction in non-convex sgd.
\newblock In \emph{Advances in Neural Information Processing Systems}, pp.\
  15210--15219, 2019.

\bibitem[Deisenroth et~al.(2013)Deisenroth, Neumann, Peters,
  et~al.]{deisenroth2013survey}
Marc~Peter Deisenroth, Gerhard Neumann, Jan Peters, et~al.
\newblock A survey on policy search for robotics.
\newblock \emph{Foundations and Trends{\textregistered} in Robotics},
  2\penalty0 (1--2):\penalty0 1--142, 2013.

\bibitem[Ding et~al.(2021)Ding, Zhang, and Lavaei]{ding2021global}
Yuhao Ding, Junzi Zhang, and Javad Lavaei.
\newblock On the global convergence of momentum-based policy gradient.
\newblock \emph{arXiv preprint arXiv:2110.10116}, 2021.

\bibitem[Fang et~al.(2018)Fang, Li, Lin, and Zhang]{fang2018spider}
Cong Fang, Chris~Junchi Li, Zhouchen Lin, and Tong Zhang.
\newblock Spider: Near-optimal non-convex optimization via stochastic
  path-integrated differential estimator.
\newblock In \emph{Advances in Neural Information Processing Systems}, pp.\
  689--699, 2018.

\bibitem[garage contributors(2019)]{garage}
The garage contributors.
\newblock Garage: A toolkit for reproducible reinforcement learning research.
\newblock \url{https://github.com/rlworkgroup/garage}, 2019.

\bibitem[Geist et~al.(2019)Geist, Scherrer, and Pietquin]{geist2019theory}
Matthieu Geist, Bruno Scherrer, and Olivier Pietquin.
\newblock A theory of regularized markov decision processes.
\newblock In \emph{Thirty-sixth International Conference on Machine Learning},
  2019.

\bibitem[Ghadimi et~al.(2016)Ghadimi, Lan, and Zhang]{ghadimi2016mini}
Saeed Ghadimi, Guanghui Lan, and Hongchao Zhang.
\newblock Mini-batch stochastic approximation methods for nonconvex stochastic
  composite optimization.
\newblock \emph{Mathematical Programming}, 155\penalty0 (1-2):\penalty0
  267--305, 2016.

\bibitem[Huang et~al.(2020)Huang, Gao, Pei, and Huang]{huang2020momentum}
Feihu Huang, Shangqian Gao, Jian Pei, and Heng Huang.
\newblock Momentum-based policy gradient methods.
\newblock In \emph{International Conference on Machine Learning}, pp.\
  4422--4433. PMLR, 2020.

\bibitem[Huang et~al.(2021)Huang, Li, and Huang]{huang2021super}
Feihu Huang, Junyi Li, and Heng Huang.
\newblock Super-adam: Faster and universal framework of adaptive gradients.
\newblock \emph{Advances in Neural Information Processing Systems}, 34, 2021.

\bibitem[Jin \& Sidford(2020)Jin and Sidford]{jin2020efficiently}
Yujia Jin and Aaron Sidford.
\newblock Efficiently solving mdps with stochastic mirror descent.
\newblock In \emph{International Conference on Machine Learning}, pp.\
  4890--4900. PMLR, 2020.

\bibitem[Johnson \& Zhang(2013)Johnson and Zhang]{johnson2013accelerating}
Rie Johnson and Tong Zhang.
\newblock Accelerating stochastic gradient descent using predictive variance
  reduction.
\newblock In \emph{NIPS}, pp.\  315--323, 2013.

\bibitem[Kakade(2001)]{kakade2001natural}
Sham~M Kakade.
\newblock A natural policy gradient.
\newblock \emph{Advances in neural information processing systems},
  14:\penalty0 1531--1538, 2001.

\bibitem[Kingma \& Ba(2014)Kingma and Ba]{kingma2014adam}
Diederik~P Kingma and Jimmy Ba.
\newblock Adam: A method for stochastic optimization.
\newblock \emph{arXiv preprint arXiv:1412.6980}, 2014.

\bibitem[Konda \& Tsitsiklis(2000)Konda and Tsitsiklis]{konda2000actor}
Vijay~R Konda and John~N Tsitsiklis.
\newblock Actor-critic algorithms.
\newblock In \emph{Advances in neural information processing systems}, pp.\
  1008--1014, 2000.

\bibitem[Lan(2021)]{lan2021policy}
Guanghui Lan.
\newblock Policy mirror descent for reinforcement learning: Linear convergence,
  new sampling complexity, and generalized problem classes.
\newblock \emph{arXiv preprint arXiv:2102.00135}, 2021.

\bibitem[Li(2017)]{li2017deep}
Yuxi Li.
\newblock Deep reinforcement learning: An overview.
\newblock \emph{arXiv preprint arXiv:1701.07274}, 2017.

\bibitem[Liu et~al.(2019)Liu, Cai, Yang, and Wang]{liu2019neural}
Boyi Liu, Qi~Cai, Zhuoran Yang, and Zhaoran Wang.
\newblock Neural proximal/trust region policy optimization attains globally
  optimal policy.
\newblock \emph{arXiv preprint arXiv:1906.10306}, 2019.

\bibitem[Liu et~al.(2020)Liu, Zhang, Basar, and Yin]{liu2020improved}
Yanli Liu, Kaiqing Zhang, Tamer Basar, and Wotao Yin.
\newblock An improved analysis of (variance-reduced) policy gradient and
  natural policy gradient methods.
\newblock \emph{Advances in Neural Information Processing Systems}, 33, 2020.

\bibitem[Neu et~al.(2017)Neu, Jonsson, and G{\'o}mez]{neu2017unified}
Gergely Neu, Anders Jonsson, and Vicen{\c{c}} G{\'o}mez.
\newblock A unified view of entropy-regularized markov decision processes.
\newblock \emph{arXiv preprint arXiv:1705.07798}, 2017.

\bibitem[Papini et~al.(2018)Papini, Binaghi, Canonaco, Pirotta, and
  Restelli]{papini2018stochastic}
Matteo Papini, Damiano Binaghi, Giuseppe Canonaco, Matteo Pirotta, and Marcello
  Restelli.
\newblock Stochastic variance-reduced policy gradient.
\newblock In \emph{35th International Conference on Machine Learning},
  volume~80, pp.\  4026--4035, 2018.

\bibitem[Paszke et~al.(2019)Paszke, Gross, Massa, Lerer, Bradbury, Chanan,
  Killeen, Lin, Gimelshein, Antiga, et~al.]{paszke2019pytorch}
Adam Paszke, Sam Gross, Francisco Massa, Adam Lerer, James Bradbury, Gregory
  Chanan, Trevor Killeen, Zeming Lin, Natalia Gimelshein, Luca Antiga, et~al.
\newblock Pytorch: An imperative style, high-performance deep learning library.
\newblock In \emph{Advances in Neural Information Processing Systems}, pp.\
  8024--8035, 2019.

\bibitem[Pham et~al.(2020)Pham, Nguyen, Phan, Nguyen, Dijk, and
  Tran-Dinh]{pham2020hybrid}
Nhan Pham, Lam Nguyen, Dzung Phan, Phuong~Ha Nguyen, Marten Dijk, and Quoc
  Tran-Dinh.
\newblock A hybrid stochastic policy gradient algorithm for reinforcement
  learning.
\newblock In \emph{International Conference on Artificial Intelligence and
  Statistics}, pp.\  374--385. PMLR, 2020.

\bibitem[Schulman et~al.(2015)Schulman, Levine, Abbeel, Jordan, and
  Moritz]{schulman2015trust}
John Schulman, Sergey Levine, Pieter Abbeel, Michael Jordan, and Philipp
  Moritz.
\newblock Trust region policy optimization.
\newblock In \emph{International conference on machine learning}, pp.\
  1889--1897, 2015.

\bibitem[Schulman et~al.(2016)Schulman, Moritz, Levine, Jordan, and
  Abbeel]{schulman2015high}
John Schulman, Philipp Moritz, Sergey Levine, Michael Jordan, and Pieter
  Abbeel.
\newblock High-dimensional continuous control using generalized advantage
  estimation.
\newblock In \emph{International Conference on Learning Representations
  (ICLR)}, 2016.

\bibitem[Schulman et~al.(2017)Schulman, Wolski, Dhariwal, Radford, and
  Klimov]{schulman2017proximal}
John Schulman, Filip Wolski, Prafulla Dhariwal, Alec Radford, and Oleg Klimov.
\newblock Proximal policy optimization algorithms.
\newblock \emph{arXiv preprint arXiv:1707.06347}, 2017.

\bibitem[Shalev-Shwartz et~al.(2016)Shalev-Shwartz, Shammah, and
  Shashua]{shalev2016safe}
Shai Shalev-Shwartz, Shaked Shammah, and Amnon Shashua.
\newblock Safe, multi-agent, reinforcement learning for autonomous driving.
\newblock \emph{arXiv preprint arXiv:1610.03295}, 2016.

\bibitem[Shani et~al.(2020)Shani, Efroni, and Mannor]{shani2020adaptive}
Lior Shani, Yonathan Efroni, and Shie Mannor.
\newblock Adaptive trust region policy optimization: Global convergence and
  faster rates for regularized mdps.
\newblock In \emph{Proceedings of the AAAI Conference on Artificial
  Intelligence}, volume~34, pp.\  5668--5675, 2020.

\bibitem[Shannon(1948)]{shannon1948mathematical}
Claude~E Shannon.
\newblock A mathematical theory of communication.
\newblock \emph{The Bell system technical journal}, 27\penalty0 (3):\penalty0
  379--423, 1948.

\bibitem[Shen et~al.(2019)Shen, Ribeiro, Hassani, Qian, and
  Mi]{shen2019hessian}
Zebang Shen, Alejandro Ribeiro, Hamed Hassani, Hui Qian, and Chao Mi.
\newblock Hessian aided policy gradient.
\newblock In \emph{International Conference on Machine Learning}, pp.\
  5729--5738, 2019.

\bibitem[Silver et~al.(2017)Silver, Schrittwieser, Simonyan, Antonoglou, Huang,
  Guez, Hubert, Baker, Lai, Bolton, et~al.]{silver2017mastering}
David Silver, Julian Schrittwieser, Karen Simonyan, Ioannis Antonoglou, Aja
  Huang, Arthur Guez, Thomas Hubert, Lucas Baker, Matthew Lai, Adrian Bolton,
  et~al.
\newblock Mastering the game of go without human knowledge.
\newblock \emph{nature}, 550\penalty0 (7676):\penalty0 354--359, 2017.

\bibitem[Sutton et~al.(2000)Sutton, McAllester, Singh, and
  Mansour]{sutton2000policy}
Richard~S Sutton, David~A McAllester, Satinder~P Singh, and Yishay Mansour.
\newblock Policy gradient methods for reinforcement learning with function
  approximation.
\newblock In \emph{Advances in neural information processing systems}, pp.\
  1057--1063, 2000.

\bibitem[Tomar et~al.(2020)Tomar, Shani, Efroni, and
  Ghavamzadeh]{tomar2020mirror}
Manan Tomar, Lior Shani, Yonathan Efroni, and Mohammad Ghavamzadeh.
\newblock Mirror descent policy optimization.
\newblock \emph{arXiv preprint arXiv:2005.09814}, 2020.

\bibitem[Tran-Dinh et~al.(2019)Tran-Dinh, Pham, Phan, and
  Nguyen]{tran2019hybrid}
Quoc Tran-Dinh, Nhan~H Pham, Dzung~T Phan, and Lam~M Nguyen.
\newblock Hybrid stochastic gradient descent algorithms for stochastic
  nonconvex optimization.
\newblock \emph{arXiv preprint arXiv:1905.05920}, 2019.

\bibitem[Wang et~al.(2019)Wang, Li, Xiong, and Zhang]{wang2019divergence}
Qing Wang, Yingru Li, Jiechao Xiong, and Tong Zhang.
\newblock Divergence-augmented policy optimization.
\newblock In \emph{Advances in Neural Information Processing Systems}, pp.\
  6099--6110, 2019.

\bibitem[Williams(1992)]{williams1992simple}
Ronald~J Williams.
\newblock Simple statistical gradient-following algorithms for connectionist
  reinforcement learning.
\newblock \emph{Machine learning}, 8\penalty0 (3-4):\penalty0 229--256, 1992.

\bibitem[Xu et~al.(2019{\natexlab{a}})Xu, Gao, and Gu]{xu2019improved}
Pan Xu, Felicia Gao, and Quanquan Gu.
\newblock An improved convergence analysis of stochastic variance-reduced
  policy gradient.
\newblock In \emph{Proceedings of the Thirty-Fifth Conference on Uncertainty in
  Artificial Intelligence}, pp.\  191, 2019{\natexlab{a}}.

\bibitem[Xu et~al.(2019{\natexlab{b}})Xu, Gao, and Gu]{xu2019sample}
Pan Xu, Felicia Gao, and Quanquan Gu.
\newblock Sample efficient policy gradient methods with recursive variance
  reduction.
\newblock \emph{arXiv preprint arXiv:1909.08610}, 2019{\natexlab{b}}.

\bibitem[Yang et~al.(2019)Yang, Zheng, Zhang, Zhang, Zheng, Wen, and
  Pan]{yang2019policy}
Long Yang, Gang Zheng, Haotian Zhang, Yu~Zhang, Qian Zheng, Jun Wen, and Gang
  Pan.
\newblock Policy optimization with stochastic mirror descent.
\newblock \emph{arXiv preprint arXiv:1906.10462}, 2019.

\bibitem[Zhan et~al.(2021)Zhan, Cen, Huang, Chen, Lee, and Chi]{zhan2021policy}
Wenhao Zhan, Shicong Cen, Baihe Huang, Yuxin Chen, Jason~D Lee, and Yuejie Chi.
\newblock Policy mirror descent for regularized reinforcement learning: A
  generalized framework with linear convergence.
\newblock \emph{arXiv preprint arXiv:2105.11066}, 2021.

\bibitem[Zhang et~al.(2021)Zhang, Ni, Yu, Szepesvari, and
  Wang]{zhang2021convergence}
Junyu Zhang, Chengzhuo Ni, Zheng Yu, Csaba Szepesvari, and Mengdi Wang.
\newblock On the convergence and sample efficiency of variance-reduced policy
  gradient method.
\newblock \emph{arXiv preprint arXiv:2102.08607}, 2021.

\bibitem[Zhang et~al.(2020)Zhang, Kim, O'Donoghue, and Boyd]{zhang2020sample}
Junzi Zhang, Jongho Kim, Brendan O'Donoghue, and Stephen Boyd.
\newblock Sample efficient reinforcement learning with reinforce.
\newblock \emph{arXiv preprint arXiv:2010.11364}, 2020.

\bibitem[Zhang et~al.(2019)Zhang, Koppel, Zhu, and
  Ba{\c{s}}ar]{zhang2019global}
Kaiqing Zhang, Alec Koppel, Hao Zhu, and Tamer Ba{\c{s}}ar.
\newblock Global convergence of policy gradient methods to (almost) locally
  optimal policies.
\newblock \emph{arXiv preprint arXiv:1906.08383}, 2019.

\bibitem[Zhang \& He(2018)Zhang and He]{zhang2018convergence}
Siqi Zhang and Niao He.
\newblock On the convergence rate of stochastic mirror descent for nonsmooth
  nonconvex optimization.
\newblock \emph{arXiv preprint arXiv:1806.04781}, 2018.

\end{thebibliography}
\bibliographystyle{iclr2022_conference}

%\appendix
%\section{Appendix}
%You may include other additional sections here.

%\appendices

\newpage

\begin{appendices}

\section{ Appendix }
\label{Appendix:A}
In this section, we study the convergence properties of our algorithms.
We first provide some useful lemmas.

\begin{lemma} \label{lem:1}
(Proposition 4.2 in \cite{xu2019sample}) Suppose $g(\tau|\theta)$ is the PGT estimator. Under Assumption 1, we have
\begin{itemize}
\item[1)] $g(\tau|\theta)$ is $L$-Lipschitz differential, i.e., $\|g(\tau|\theta_1)-g(\tau|\theta_2)\|\leq L\|\theta_1-\theta_2\|$  for all
$\theta_1, \theta_2 \in \Theta$, where $L=C_hR/(1-\gamma)^2$;
\item[2)] $J(\theta)$ is $L$-smooth, i.e., $\|\nabla^2 J(\theta)\|\leq L$;
\item[3)] $g(\tau|\theta)$ is bounded, i.e., $\|g(\tau|\theta)\|\leq G$ for all $\theta\in \Theta$ with $G=C_gR/(1-\gamma)^2$.
\end{itemize}
\end{lemma}
\begin{lemma} \label{lem:2}
(Lemma 6.1 in \cite{xu2019improved}) Under Assumptions 1 and 3, let $w(\tau|\theta_{k-1},\theta_k)= g(\tau|\theta_{k-1})/g(\tau | \theta_k)$, we have
\begin{align}
\mathbb{V}[w(\tau|\theta_{k-1},\theta_k)] \leq C^2_w\|\theta_k-\theta_{k-1}\|^2,
\end{align}
where $C_w = \sqrt{H(2HC_g^2+C_h)(W+1)}$.
\end{lemma}

\begin{lemma} \label{lem:A1}
(Lemma 1 in \citep{ghadimi2016mini})
Let $\mathcal{X} \subseteq \mathbb{R}^d$ be a closed convex set, and $\phi: \mathcal{X} \rightarrow \mathbb{R}$ be a convex function
but possibly nonsmooth, and $D_{\psi}: \mathcal{X}\times \mathcal{X} \rightarrow \mathbb{R}$ is Bregman
divergence related to the $\nu$-strongly convex function $\psi$. Then we define
\begin{align}
  & x^+ = \arg\min_{z\in \mathcal{X}} \big\{ \langle g, z \rangle + \frac{1}{\lambda}D_{\psi}(z,x) + \phi(z)\big\}, \ \forall x\in \mathcal{X}  \label{eq:A17} \\
  & P_{\mathcal{X}}(x,g,\lambda) = \frac{1}{\lambda} (x-x^+), \label{eq:A18}
\end{align}
where $g\in \mathbb{R}^d$, $\lambda >0$ and $D_{\psi}(z,x) = \psi(z) - \big(\psi(x) + \langle\nabla\psi(x),z-x\rangle\big)$.
Then the following statement holds
\begin{align}
  \langle g, P_{\mathcal{X}}(x,g,\lambda)\rangle \geq \nu \|P_{\mathcal{X}}(x,g,\lambda)\|^2 + \frac{1}{\lambda}\big[\phi(x^+)-\phi(x)\big].
\end{align}
\end{lemma}

\begin{lemma} \label{lem:A2}
(Proposition 1 in \cite{ghadimi2016mini})
Let $x_1^+$ and $x_2^+$ be given in (\ref{eq:A17}) with $g$ replaced by $g_1$ and $g_2$ respectively. Then let
$P_{\mathcal{X}}(x,g_1,\lambda)$ and $P_{\mathcal{X}}(x,g_2,\lambda)$ be defined in (\ref{eq:A18}) with $x^+$ replaced by $x_1^+$ and $x_2^+$ respectively.
we have
\begin{align}
  \|P_{\mathcal{X}}(x,g_1,\lambda)-P_{\mathcal{X}}(x,g_2,\lambda)\| \leq \frac{1}{\nu}\|g_1-g_2\|.
\end{align}
\end{lemma}

\begin{lemma} \label{lem:A3}
(Lemma 1 in \citep{cortes2010learning}) Let $w(x)=\frac{P(x)}{Q(x)}$ be the importance weight for distributions $P$ and $Q$.
The following identities hold for the expectation, second moment, and variance of $w(x)$
 \begin{align}
 & \mathbb{E}[w(x)] = 1, \ \mathbb{E}[w^2(x)]= d_2(P||Q), \nonumber \\
 & \mathbb{V}[w(x)] = d_2(P||Q)-1,
 \end{align}
where $d_2(P||Q)=2^{D(P||Q)}$, and $D(P||Q)$ is  $R\acute{e}nyi$ divergence between distributions $P$ and $Q$.
\end{lemma}

 \begin{lemma} \label{lem:B1}
 Suppose that the sequence $\{\theta_k\}_{k=1}^K$ be generated from Algorithms \ref{alg:1} or \ref{alg:2}.
 Let $0<\eta_k \leq 1$ and $0< \lambda \leq \frac{\nu}{2L\eta_k}$, then we have
 \begin{align}
 f(\theta_{k+1}) - f(\theta_k) \leq \frac{\eta_k\lambda}{\nu}\|\nabla f(\theta_k)-u_k\|^2-\frac{\nu\eta_k}{2\lambda}\|\tilde{\theta}_{k+1}-\theta_k\|^2.
 \end{align}
 \end{lemma}

 \begin{proof}
 According to Assumption 1 and Lemma \ref{lem:1}, the function $f(\theta)$ is $L$-smooth.
 Then we have
 \begin{align} \label{eq:B1}
 f(\theta_{k+1}) & \leq f(\theta_k) + \langle\nabla f(\theta_k), \theta_{k+1}-\theta_k\rangle + \frac{L}{2}\|\theta_{k+1}-\theta_k\|^2 \\
 & = f(\theta_k) + \eta_k\langle \nabla f(\theta_k),\tilde{\theta}_{k+1}-\theta_k\rangle + \frac{L\eta_k^2}{2}\|\tilde{\theta}_{k+1}-\theta_k\|^2 \nonumber \\
 & = f(\theta_k) + \eta_k\langle \nabla f(\theta_k)-u_k,\tilde{\theta}_{k+1}-\theta_k\rangle + \eta_k\langle u_k,\tilde{\theta}_{k+1}-\theta_k\rangle + \frac{L\eta_k^2}{2}\|\tilde{\theta}_{k+1}-\theta_k\|^2, \nonumber
 \end{align}
 where the second equality is due to $\theta_{k+1}=\theta_k + \eta_k(\tilde{\theta}_{k+1}-\theta_k)$.
 By the step 4 of Algorithm \ref{alg:1} or \ref{alg:2}, we have $\tilde{\theta}_{k+1} = \arg\min_{\theta\in \Theta} \big\{ \langle u_k, \theta\rangle + \frac{1}{\lambda} D_{\psi_k}(\theta,\theta_k) \big\}$.
 By using Lemma \ref{lem:A1} with $\phi(\cdot)=0$, we have
 \begin{align}
 \langle u_k, \frac{1}{\lambda}(\theta_k - \tilde{\theta}_{k+1})\rangle \geq \nu\|\frac{1}{\lambda}(\theta_k - \tilde{\theta}_{k+1})\|^2.
 \end{align}
 Thus, we can obtain
 \begin{align} \label{eq:B3}
 \langle u_k, \tilde{\theta}_{k+1}-\theta_k\rangle \leq -\frac{\nu }{\lambda}\|\tilde{\theta}_{k+1}-\theta_k\|^2.
 \end{align}
 According to the Cauchy-Schwarz inequality and Young's inequality, we have
 \begin{align} \label{eq:B4}
 \langle \nabla f(\theta_k)-u_k,\tilde{\theta}_{k+1}-\theta_k\rangle & \leq \|\nabla f(\theta_k)-u_k\|\|\tilde{\theta}_{k+1}-\theta_k\| \nonumber \\
 & \leq \frac{\lambda}{\nu}\|\nabla f(\theta_k)-u_k\|^2+\frac{\nu}{4\lambda}\|\tilde{\theta}_{k+1}-\theta_k\|^2.
 \end{align}
 Combining the inequalities (\ref{eq:B1}), (\ref{eq:B3}) with (\ref{eq:B4}),
 we obtain
 \begin{align}
 f(\theta_{k+1}) & \leq f(\theta_k) + \eta_k\langle \nabla f(\theta_k)-u_k,\tilde{\theta}_{k+1}-\theta_k\rangle + \eta_k\langle u_k,\tilde{\theta}_{k+1}-\theta_k\rangle + \frac{L\eta_k^2}{2}\|\tilde{\theta}_{k+1}-\theta_k\|^2 \nonumber \\
 & \leq f(\theta_k) + \frac{\eta_k\lambda}{\nu}\|\nabla f(\theta_k)-u_k\|^2 + \frac{\nu\eta_k}{4\lambda}\|\tilde{\theta}_{k+1}-\theta_k\|^2 -\frac{\nu\eta_k}{\lambda}\|\tilde{\theta}_{k+1}-\theta_k\|^2 + \frac{L\eta_k^2}{2}\|\tilde{\theta}_{k+1}-\theta_k\|^2 \nonumber \\
 & = f(\theta_k) + \frac{\eta_k\lambda}{\nu}\|\nabla f(\theta_k)-u_k\|^2 -\frac{\nu\eta_k}{2\lambda}\|\tilde{\theta}_{k+1}-\theta_k\|^2 -\big(\frac{\nu\eta_k}{4\lambda}-\frac{L\eta_k^2}{2}\big)\|\tilde{\theta}_{k+1}-\theta_k\|^2 \nonumber \\
 & \leq f(\theta_k) + \frac{\eta_k\lambda}{\nu}\|\nabla f(\theta_k)-u_k\|^2 -\frac{\nu\eta_k}{2\lambda}\|\tilde{\theta}_{k+1}-\theta_k\|^2,
 \end{align}
 where the last inequality is due to $0< \lambda \leq \frac{\nu}{2L\eta_k}$.
 \end{proof}

 \subsection{ Convergence Analysis of BGPO Algorithm }
 \label{Appendix:A1}
 In this subsection, we analyze the convergence properties of BGPO algorithm.

 \begin{lemma} \label{lem:B2}
 Assume the stochastic policy gradient $u_{k+1}$ be generated from Algorithm \ref{alg:1}, given $0<\beta_k\leq 1$,
 we have
 \begin{align}
  \mathbb{E}\|\nabla f(\theta_{k+1})-u_{k+1}\|^2 \leq (1-\beta_{k+1})\mathbb{E} \|\nabla f(\theta_k) - u_k\|^2 + \frac{2}{\beta_{k+1}}L^2\eta_k^2\mathbb{E}\| \tilde{\theta}_{k+1} - \theta_k\|^2  + \beta_{k+1}^2\sigma^2. \nonumber
 \end{align}
 \end{lemma}
 \begin{proof}
 By the definition of $u_{k+1}$ in Algorithm \ref{alg:1}, we have
  \begin{align}
  u_{k+1} - u_k = -\beta_{k+1}u_k - \beta_{k+1} g(\tau_{k+1}|\theta_{k+1}).
  \end{align}
 Since $\nabla f(\theta_k)=-J(\theta_k)$ for all $k\geq 1$, we have
  \begin{align}
  & \mathbb{E}\|\nabla f(\theta_{k+1}) - u_{k+1}\|^2 \nonumber \\
  &= \mathbb{E}\|-\nabla J(\theta_k) - u_k - \nabla J(\theta_{k+1}) + \nabla J(\theta_k) - (u_{k+1}-u_k)\|^2\nonumber \\
  & = \mathbb{E}\|-\nabla J(\theta_k) - u_k - \nabla J(\theta_{k+1}) + \nabla J(\theta_k) + \beta_{k+1}u_k + \beta_{k+1}g(\tau_{k+1}|\theta_{k+1})\|^2\nonumber \\
  & = \mathbb{E}\|(1-\beta_{k+1})(-\nabla J(\theta_k) - u_k) + \beta_{k+1}(-\nabla J(\theta_{k+1})+g(\tau_{k+1}|\theta_{k+1})) \nonumber \\
  & \quad + (1-\beta_{k+1})\big( -\nabla J(\theta_{k+1})+\nabla J(\theta_k)\big)\|^2\nonumber \\
  & = (1-\beta_{k+1})^2\mathbb{E} \|\nabla J(\theta_k) + u_k + \nabla J(\theta_{k+1})-\nabla J(\theta_k)\|^2  + \beta_{k+1}^2\mathbb{E} \|\nabla J(\theta_{k+1})-g(\tau_{k+1}|\theta_{k+1}) \|^2 \nonumber \\
  & \leq (1-\beta_{k+1})^2(1+\beta_{k+1})\mathbb{E} \|\nabla J(\theta_k) + u_k\|^2 + (1-\beta_{k+1})^2(1+\frac{1}{\beta_{k+1}})\mathbb{E}\|\nabla J(\theta_{k+1}) - \nabla J(\theta_k)\|^2  \nonumber \\
  & \quad + \beta_{k+1}^2\mathbb{E} \|\nabla J(\theta_{k+1}) - g(\tau_{k+1}|\theta_{k+1}) \|^2 \nonumber \\
  & \leq (1-\beta_{k+1})\mathbb{E} \|\nabla J(\theta_k) + u_k\|^2 + \frac{2}{\beta_{k+1}}\mathbb{E}\|\nabla J(\theta_{k+1})-\nabla J(\theta_k)\|^2  + \beta_{k+1}^2\mathbb{E} \|\nabla J(\theta_{k+1})-g(\tau_{k+1}|\theta_{k+1}) \|^2 \nonumber \\
  & \leq  (1-\beta_{k+1})\mathbb{E} \|\nabla J(\theta_k) + u_k\|^2 + \frac{2}{\beta_{k+1}}L^2\mathbb{E}\| \theta_{k+1} - \theta_k\|^2  + \beta_{k+1}^2\mathbb{E} \|\nabla J(\theta_{k+1})-g(\tau_{k+1}|\theta_{k+1}) \|^2 \nonumber \\
  & \leq (1-\beta_{k+1})\mathbb{E} \|\nabla f(\theta_k) - u_k\|^2 + \frac{2}{\beta_{k+1}}L^2\eta_k^2\mathbb{E}\| \tilde{\theta}_{k+1} - \theta_k\|^2  + \beta_{k+1}^2\sigma^2,
  \end{align}
 where the fourth equality holds by $\mathbb{E}_{\tau_{k+1}\sim p(\tau|\theta_{k+1})}[g(\tau_{k+1}|\theta_{k+1})]=\nabla J(\theta_{k+1})$;
 the first inequality holds by Young's inequality; the second inequality is due to $0< \beta_{k+1} \leq 1$ such that  $(1-\beta_{k+1})^2(1+\beta_{k+1})=1-\beta_{k+1}-\beta_{k+1}^2+
  \beta_{k+1}^3\leq 1-\beta_{k+1}$ and $(1-\beta_{k+1})^2(1+\frac{1}{\beta_{k+1}}) \leq 1+\frac{1}{\beta_{k+1}} \leq \frac{2}{\beta_{k+1}}$;
 the last inequality holds by Assumption 2.
 \end{proof}

 \begin{theorem} \label{th:A1}
 Assume the sequence $\{\theta_k\}_{k=1}^K$ be generated from Algorithm \ref{alg:1}.
 Let $\eta_k=\frac{b}{(m+k)^{1/2}}$ for all $k\geq 1$, $0< \lambda \leq \frac{\nu m^{1/2}}{9Lb}$, $b>0$, $\frac{8L\lambda}{\nu} \leq c \leq \frac{m^{1/2}}{b}$,
 and $m \geq \max\{b^2,(cb)^2\}$, we have
 \begin{align}
  \frac{1}{K} \sum_{k=1}^K \mathbb{E} \|\mathcal{B}^{\psi_k}_{\lambda,\langle \nabla f(\theta_k), \theta\rangle}(\theta_k)\|
  \leq \frac{2\sqrt{2M}m^{1/4}}{K^{1/2}} + \frac{2\sqrt{2M}}{K^{1/4}}, \nonumber
 \end{align}
 where $M = \frac{J^* - J(\theta_1)}{\nu\lambda b} + \frac{\sigma^2}{\nu\lambda Lb} + \frac{m\sigma^2}{\nu \lambda Lb}\ln(m+K)$.
 \end{theorem}

 \begin{proof}
 Since $\eta_k=\frac{b}{(m+k)^{1/2}}$ is decreasing on $k$, we have $\eta_k \leq \eta_0 = \frac{b}{m^{1/2}}\leq 1$ for all $k\geq 0$.
 At the same time, let $m\geq (cb)^2$,
 we have $\beta_{k+1} = c\eta_k \leq c\eta_0 = \frac{cb}{m^{1/2}} \leq 1$. Consider $m\geq (cb)^2$, we have $c\leq \frac{m^{1/2}}{b}$.
 Since $0<\lambda \leq \frac{\nu m^{1/2}}{9Lb}$, we have $\lambda \leq \frac{\nu m^{1/2}}{9Lb}\leq \frac{\nu m^{1/2}}{2Lb} = \frac{\nu}{2L \eta_0}\leq \frac{\nu}{2L \eta_k}$ for all $k\geq 0$.
 According to Lemma \ref{lem:B2}, we have
 \begin{align} \label{eq:E2}
  & \mathbb{E}\|\nabla f(\theta_{k+1})-u_{k+1}\|^2 - \mathbb{E} \|\nabla f(\theta_k) - u_k\|^2 \nonumber \\
  & \leq -\beta_{k+1}\mathbb{E} \|\nabla f(\theta_k) - u_k\|^2 + \frac{2}{\beta_{k+1}}L^2\eta_k^2\mathbb{E}\| \tilde{\theta}_{k+1} - \theta_k\|^2  + \beta_{k+1}^2\sigma^2 \nonumber \\
  & = -c\eta_k\mathbb{E} \|\nabla f(\theta_k) - u_k\|^2 + \frac{2 L^2}{c}\eta_k\mathbb{E}\| \tilde{\theta}_{k+1} - \theta_k\|^2  + c^2\eta_k^2\sigma^2 \nonumber \\
  & = -\frac{8L\lambda}{\nu}\eta_k\mathbb{E} \|\nabla f(\theta_k) - u_k\|^2 + \frac{L\nu}{4\lambda}\eta_k\mathbb{E}\| \tilde{\theta}_{k+1} - \theta_k\|^2  + \frac{m\eta_k^2\sigma^2}{b^2},
 \end{align}
 where the first equality is due to $\beta_{k+1} = c\eta_k$ and the last equality holds by $\frac{8L\lambda}{\nu} \leq c \leq \frac{m^{1/2}}{b}$.

 Next we define a \emph{Lyapunov} function $\Phi_k = \mathbb{E}\big[f(\theta_k) + \frac{1}{L}\|\nabla f(\theta_k)-u_k\|^2\big]$ for any $t\geq 1$.
 Then we have
  \begin{align} \label{eq:E3}
  \Phi_{k+1} - \Phi_k & = \mathbb{E}\big[ f(\theta_{k+1}) - f(\theta_k) + \frac{1}{L}\big( \|\nabla f(\theta_{k+1})-u_{k+1}\|^2 - \|\nabla f(\theta_k)-u_k\|^2 \big) \big] \nonumber \\
  & \leq \frac{\lambda\eta_k}{\nu}\mathbb{E}\|\nabla f(\theta_k)-u_k\|^2 - \frac{\nu\eta_k}{2\lambda}\mathbb{E}\|\tilde{\theta}_{k+1}-\theta_k\|^2 -\frac{8\lambda\eta_k}{\nu}\mathbb{E} \|\nabla f(\theta_k) - u_k\|^2
  \nonumber \\
  & \quad + \frac{\nu\eta_k}{4\lambda}\mathbb{E}\| \tilde{\theta}_{k+1} - \theta_k\|^2 + \frac{m\eta_k^2\sigma^2}{Lb^2} \nonumber \\
  & \leq -\frac{\lambda}{4\nu}\eta_k\mathbb{E}\|\nabla f(\theta_k)-u_k\|^2 - \frac{\nu}{4\lambda}\eta_k\mathbb{E}\| \tilde{\theta}_{k+1} - \theta_k\|^2
  + \frac{m\eta_k^2\sigma^2}{Lb^2},
  \end{align}
 where the first inequality follows by the Lemma \ref{lem:B1} and the above inequality (\ref{eq:E2}).

 Summing the above inequality (\ref{eq:E3}) over $k$ from $1$ to $K$, we can obtain
 \begin{align} \label{eq:E4}
  & \sum_{k=1}^K\mathbb{E}[\frac{\lambda}{4\nu}\eta_k\mathbb{E}\|\nabla f(\theta_k)-u_k\|^2 + \frac{\nu}{4\lambda}\eta_k\mathbb{E}\| \tilde{\theta}_{k+1} - \theta_k\|^2] \nonumber \\
  & \leq \Phi_1 - \Phi_{K+1} + \frac{m\sigma^2}{Lb^2}\sum_{k=1}^K\eta_k^2 \nonumber \\
  & = f(\theta_1) - f(\theta_{K+1}) + \frac{1}{L}\|\nabla f(\theta_1)-u_1\|^2 - \frac{1}{L}\|\nabla f(\theta_{K+1})-u_{K+1}\|^2+ \frac{m\sigma^2}{Lb^2}\sum_{k=1}^K \eta_k^2 \nonumber \\
  & \leq J(\theta_{K+1})-J(\theta_1) + \frac{1}{L}\|\nabla J(\theta_1)-g(\tau_1|\theta_1)\|^2
  + \frac{m\sigma^2}{Lb^2}\sum_{k=1}^K \eta_k^2 \nonumber \\
  & \leq J^* - J(\theta_1) + \frac{\sigma^2}{L} + \frac{m\sigma^2}{L}\int_{1}^K\frac{1}{m+k} dk \nonumber \\
  & \leq J^* - J(\theta_1) + \frac{\sigma^2}{L} + \frac{m\sigma^2}{L}\ln(m+K),
 \end{align}
 where the last second inequality holds by Assumptions 2 and 4.

 Since $\eta_k$ is decreasing, we have
 \begin{align} \label{eq:E5}
  & \frac{1}{K}\sum_{k=1}^K\mathbb{E}[\frac{1}{4\nu^2}\mathbb{E}\|\nabla f(\theta_k)-u_k\|^2 + \frac{1}{4\lambda^2}\mathbb{E}\| \tilde{\theta}_{k+1} - \theta_k\|^2] \nonumber \\
  & \leq  \frac{J^* - J(\theta_1)}{K\nu\lambda\eta_K} + \frac{\sigma^2}{K\nu\lambda L\eta_K} + \frac{m\sigma^2}{\nu \lambda LK\eta_K}
  \ln(m+K) \nonumber \\
  & \leq \bigg( \frac{J^* - J(\theta_1)}{\nu\lambda b} + \frac{\sigma^2}{\nu\lambda Lb} + \frac{m\sigma^2}{\nu \lambda Lb}\ln(m+K) \bigg)\frac{(m+K)^{1/2}}{K}.
 \end{align}
 Let $M = \frac{J^* - J(\theta_1)}{\nu\lambda b} + \frac{\sigma^2}{\nu\lambda Lb} + \frac{m\sigma^2}{\nu \lambda Lb}\ln(m+K)$,
 the above inequality (\ref{eq:E5}) reduces to
 \begin{align}
  \frac{1}{K}\sum_{k=1}^K\mathbb{E}[\frac{1}{4\nu^2}\mathbb{E}\|\nabla f(\theta_k)-u_k\|^2 + \frac{1}{4\lambda^2}
  \mathbb{E}\| \tilde{\theta}_{k+1} - \theta_k\|^2] \leq \frac{M}{K}(m+K)^{1/2}.
 \end{align}

 According to Jensen's inequality, we have
 \begin{align}
  & \frac{1}{K} \sum_{k=1}^K\mathbb{E} \big[\frac{1}{2\nu} \|\nabla f(\theta_k) -u_k\|
  + \frac{1}{2\lambda} \|\tilde{\theta}_{k+1}-\theta_k\| \big] \nonumber \\
  & \leq \big( \frac{2}{K} \sum_{k=1}^K\mathbb{E} \big[\frac{1}{4\nu^2} \|\nabla f(\theta_k) -u_k\|^2
  + \frac{1}{4\lambda^2} \|\tilde{\theta}_{k+1}-\theta_k\|^2\big]\big)^{1/2} \nonumber \\
  & \leq \frac{\sqrt{2M}}{K^{1/2}}(m+K)^{1/4} \leq \frac{\sqrt{2M}m^{1/4}}{K^{1/2}} + \frac{\sqrt{2M}}{K^{1/4}},
 \end{align}
 where the last inequality is due to the inequality $(a+b)^{1/4} \leq a^{1/4} + b^{1/4}$ for all $a,b\geq0$.
 Thus we have
 \begin{align} \label{eq:E6}
  \frac{1}{K} \sum_{k=1}^K\mathbb{E} \big[ \frac{1}{\nu}\|\nabla f(\theta_k) -u_k\| + \frac{1}{\lambda} \|\tilde{\theta}_{k+1}-\theta_k\| \big]
  \leq \frac{2\sqrt{2M}m^{1/4}}{K^{1/2}} + \frac{2\sqrt{2M}}{K^{1/4}}.
 \end{align}

 By the step 4 of Algorithm \ref{alg:1}, we have
  \begin{align}
  \mathcal{B}^{\psi_k}_{\lambda,\langle u_k, \theta\rangle}(\theta_k) = P_{\Theta}(\theta_k,u_k,\lambda) = \frac{1}{\lambda}\big(\theta_k - \tilde{\theta}_{k+1}\big).
 \end{align}
 At the same time, as in \cite{ghadimi2016mini}, we define
 \begin{align}
 \mathcal{B}^{\psi_k}_{\lambda,\langle \nabla f(\theta_k), \theta\rangle}(\theta_k) = P_{\Theta}(\theta_k,\nabla f(\theta_k),\lambda) = \frac{1}{\lambda}\big(\theta_k - \theta^+_{k+1}\big),
 \end{align}
 where
 \begin{align}
  \theta^+_{k+1} = \arg\min_{\theta\in \Theta} \big\{ \langle \nabla f(\theta_k), \theta\rangle
+ \frac{1}{\lambda} D_{\psi_k}(\theta,\theta_k) \big\}.
 \end{align}
According to the above Lemma \ref{lem:A2}, we have
$\|\mathcal{B}^{\psi_k}_{\lambda,\langle u_k, \theta\rangle}(\theta_k) - \mathcal{B}^{\psi_k}_{\lambda,\langle \nabla f(\theta_k), \theta\rangle}(\theta_k)\| \leq \frac{1}{\nu} \|u_k - \nabla f(\theta_k)\|$.
Then we have
\begin{align} \label{eq:E7}
 \|\mathcal{B}^{\psi_k}_{\lambda,\langle \nabla f(\theta_k), \theta\rangle}(\theta_k)\| & \leq
 \|\mathcal{B}^{\psi_k}_{\lambda,\langle u_k, \theta\rangle}(\theta_k)\| +\|\mathcal{B}^{\psi_k}_{\lambda,\langle u_k, \theta\rangle}(\theta_k) - \mathcal{B}^{\psi_k}_{\lambda,\langle \nabla f(\theta_k), \theta\rangle}(\theta_k)\| \nonumber \\
 & \leq \|\mathcal{B}^{\psi_k}_{\lambda,\langle u_k, \theta\rangle}(\theta_k)\| + \frac{1}{\nu} \|u_k - \nabla f(\theta_k)\| \nonumber \\
 & = \frac{1}{\lambda} \|\tilde{\theta}_{k+1}-\theta_k\| + \frac{1}{\nu} \|u_k - \nabla f(\theta_k)\|.
\end{align}

By the above inequalities (\ref{eq:E6}) and (\ref{eq:E7}), we have
\begin{align}
 \frac{1}{K} \sum_{k=1}^K \mathbb{E} \|\mathcal{B}^{\psi_k}_{\lambda,\langle \nabla f(\theta_k), \theta\rangle}(\theta_k)\|
  \leq \frac{2\sqrt{2M}m^{1/4}}{K^{1/2}} + \frac{2\sqrt{2M}}{K^{1/4}}.
\end{align}

 \end{proof}

 \subsection{ Convergence Analysis of VR-BGPO algorithm}
 \label{Appendix:A2}
 In this subsection, we will analyze convergence properties of the VR-BGPO algorithm.

\begin{lemma} \label{lem:C1}
Assume that the stochastic policy gradient $u_{k+1}$ be generated from Algorithm \ref{alg:2}, given $0<\beta_k\leq 1$, we have
 \begin{align}
 \mathbb{E}\|\nabla f(\theta_{k+1})-u_{k+1}\|^2 \leq (1-\beta_{k+1})\mathbb{E} \|\nabla f(\theta_k)-u_k\|^2
 + 4\hat{L}^2\eta^2_k\|\tilde{\theta}_{k+1}-\theta_k\|^2 + 2\beta^2_{k+1}\sigma^2, \nonumber
 \end{align}
where $\hat{L}^2 = L^2 + 2G^2C^2_w$ and $C_w = \sqrt{H(2HC_g^2+C_h)(W+1)}$.
\end{lemma}

 \begin{proof}
 By the definition of $u_{k+1}$ in Algorithm \ref{alg:2}, we have
 \begin{align}
  & u_{k+1} - u_k \nonumber \\
  & = -\beta_{k+1}u_k - \beta_{k+1} g(\tau_{k+1}|\theta_{k+1}) + (1-\beta_{k+1})\big( -g(\tau_{k+1} | \theta_{k+1})+ w(\tau_{k+1}|\theta_k,\theta_{k+1}) g(\tau_{k+1}|\theta_k)\big). \nonumber
 \end{align}
 Since $\nabla f(\theta_{k+1})=-\nabla J(\theta_{k+1})$, we have
 \begin{align} \label{eq:C1}
  &\mathbb{E}\|\nabla f(\theta_{k+1})-u_{k+1}\|^2\nonumber \\
  & = \mathbb{E}\|\nabla J(\theta_k) + u_k + \nabla J(\theta_{k+1})-\nabla J(\theta_k) +(u_{k+1}-u_k)\|^2 \\
  & =  \mathbb{E}\|\nabla J(\theta_k) + u_k + \nabla J(\theta_{k+1})-\nabla J(\theta_k) - \beta_{k+1}u_k- \beta_{k+1} g(\tau_{k+1}|\theta_{k+1})\nonumber \\
  & \quad + (1-\beta_{k+1})\big( -g(\tau_{k+1} | \theta_{k+1})+ w(\tau_{k+1}|\theta_k,\theta_{k+1}) g(\tau_{k+1}|\theta_k)\big)\|^2 \nonumber \\
  & = \mathbb{E}\|(1-\beta_{k+1})(\nabla J(\theta_k) + u_k) + \beta_{k+1}(\nabla J(\theta_{k+1})- g(\tau_{k+1}|\theta_{k+1}))\nonumber \\
  & \quad - (1-\beta_{k+1})\big( g(\tau_{k+1} | \theta_{k+1})- w(\tau_{k+1}|\theta_k,\theta_{k+1}) g(\tau_{k+1}|\theta_k)-(\nabla J(\theta_{k+1})-\nabla J(\theta_k))\big)\|^2\nonumber \\
  & =(1-\beta_{k+1})^2\mathbb{E}\|\nabla J(\theta_k) + u_k\|^2 + \mathbb{E}\|\beta_{k+1}(\nabla J(\theta_{k+1})
  - g(\tau_{k+1}|\theta_{k+1}))\nonumber \\
  & \quad - (1-\beta_{k+1})\big( g(\tau_{k+1} | \theta_{k+1})- w(\tau_{k+1}|\theta_k,\theta_{k+1}) g(\tau_{k+1}|\theta_k)-(\nabla J(\theta_{k+1})-\nabla J(\theta_k))\big)\|^2\nonumber \\
  & \leq (1-\beta_{k+1})^2\mathbb{E} \|\nabla J(\theta_k) + u_k\|^2 + 2\beta^2_{k+1}\mathbb{E}\|\nabla J(\theta_{k+1})- g(\tau_{k+1}|\theta_{k+1})\|^2\nonumber \\
  & \quad +2(1-\beta_{k+1})^2 \mathbb{E}\| g(\tau_{k+1} | \theta_{k+1})- w(\tau_{k+1}|\theta_k,\theta_{k+1}) g(\tau_{k+1}|\theta_k)-(\nabla J(\theta_{k+1})-\nabla J(\theta_k))\|^2 \nonumber \\
  & \leq (1-\beta_{k+1})\mathbb{E} \|\nabla f(\theta_k) - u_k\|^2 + 2\beta^2_{k+1}\sigma^2 + 2
  \underbrace{ \mathbb{E}\| g(\tau_{k+1} | \theta_{k+1})- w(\tau_{k+1}|\theta_k,\theta_{k+1}) g(\tau_{k+1}|\theta_k)\|^2 }_{=T_1}, \nonumber
 \end{align}
 where the forth equality holds by $\mathbb{E}_{\tau_{k+1}\sim p(\tau|\theta_{k+1})}[g(\tau_{k+1}|\theta_{k+1})]=\nabla J(\theta_{k+1})$ and $\mathbb{E}_{\tau_{k+1}\sim p(\tau|\theta_{k+1})}[g(\tau_{k+1}|\theta_{k+1})- w(\tau_{k+1}|\theta_k,\theta_{k+1}) g(\tau_{k+1}|\theta_k)]=\nabla J(\theta_{k+1})-\nabla J(\theta_k)$; the second last inequality follows by Young's inequality; and the last inequality holds by Assumption 2,
 and the inequality $\mathbb{E}\|\zeta-\mathbb{E}[\zeta]\|^2=\mathbb{E}\|\zeta\|^2-(\mathbb{E}[\zeta])^2 \leq \mathbb{E}\|\zeta\|^2$, and
 $0<\beta_{k+1}\leq 1$.

 Next, we give an upper bound of the term $T_1$ as follows:
  \begin{align} \label{eq:C2}
  T_1 & = \mathbb{E}\| g(\tau_{k+1} | \theta_{k+1})- w(\tau_{k+1}|\theta_k,\theta_{k+1}) g(\tau_{k+1}|\theta_k)\|^2 \nonumber \\
  & = \mathbb{E}\| g(\tau_{k+1} | \theta_{k+1})-g(\tau_{k+1}|\theta_k) + g(\tau_{k+1}|\theta_k)- w(\tau_{k+1}|\theta_k,\theta_{k+1}) g(\tau_{k+1}|\theta_k)\|^2 \nonumber \\
  & \leq 2\mathbb{E}\| g(\tau_{k+1} | \theta_{k+1})-g(\tau_{k+1}|\theta_k)\|^2 + 2\mathbb{E}\|(1- w(\tau_{k+1}|\theta_k,\theta_{k+1})) g(\tau_{k+1}|\theta_k)\|^2 \nonumber \\
  & \leq 2L^2\|\theta_{k+1} - \theta_k\|^2 + 2G^2\mathbb{E}\|1- w(\tau_{k+1}|\theta_k,\theta_{k+1})\|^2 \nonumber \\
  & = 2L^2\|\theta_{k+1} - \theta_k\|^2 + 2G^2\mathbb{V}\big(w(\tau_{k+1}|\theta_k,\theta_{k+1})\big) \nonumber \\
  & \leq  2(L^2 + 2G^2C^2_w)\|\theta_{k+1}-\theta_k\|^2,
  \end{align}
 where the second inequality holds by Lemma \ref{lem:1}, and the third equality holds by Lemma \ref{lem:A3},
 and the last inequality follows by Lemma \ref{lem:2}.

 Combining the inequalities (\ref{eq:C1}) with (\ref{eq:C2}), let $\hat{L}^2 = L^2 + 2G^2C^2_w$, we have
 \begin{align}
 \mathbb{E}\|\nabla f(\theta_{k+1})-u_{k+1}\|^2& \leq (1-\beta_{k+1})\mathbb{E} \|\nabla f(\theta_k)-u_k\|^2 + 2\beta^2_{k+1}\sigma^2+ 4\hat{L}^2\|\theta_{k+1}-\theta_k\|^2 \nonumber \\
 & = (1-\beta_{k+1})\mathbb{E}\|\nabla f(\theta_k)-u_k\|^2 + 2\beta^2_{k+1}\sigma^2
 + 4\hat{L}^2\eta^2_k\|\tilde{\theta}_{k+1}-\theta_k\|^2. \nonumber
 \end{align}

 \end{proof}

 \begin{theorem} \label{th:A2}
 Suppose the sequence $\{\theta_k\}_{k=1}^K$ be generated from Algorithm \ref{alg:2}. Let $\eta_k = \frac{b}{(m+k)^{1/3}}$
 for all $k \geq 0$, $0< \lambda \leq \frac{\nu m^{1/3}}{5\hat{L}b}$, $b>0$,
 $\frac{2}{3b^3} + \frac{20\hat{L}^2 \lambda^2}{\nu^2} \leq c \leq \frac{m^{2/3}}{b^2}$ and $m\geq \max\big(2,b^3,(cb)^3,(\frac{5}{6b})^{2/3}\big)$,
 we have
 \begin{align}
  \frac{1}{K} \sum_{k=1}^K \mathbb{E} \|\mathcal{B}^{\psi_k}_{\lambda,\langle \nabla f(\theta_k), \theta\rangle}(\theta_k)\|
  \leq \frac{2\sqrt{2M'}m^{1/6}}{K^{1/2}} + \frac{2\sqrt{2M'}}{K^{1/3}},
 \end{align}
 where $M'=\frac{J^*-J(\theta_1)}{b\nu\lambda} + \frac{m^{1/3}\sigma^2}{16b^2\hat{L}^2\lambda^2}
  + \frac{c^2\sigma^2 b^2}{8\hat{L}^2\lambda^2}$ and $\hat{L}^2 = L^2 + 2G^2C^2_w$.
 \end{theorem}

 \begin{proof}
 Since $\eta_k=\frac{b}{(m+k)^{1/3}}$ on $k$ is decreasing and $m\geq b^3$, we have $\eta_k \leq \eta_0 = \frac{b}{m^{1/3}} \leq 1$.
 Due to $\hat{L} = \sqrt{L^2 + 2G^2C^2_w} \geq L$, we have $0<\lambda \leq  \frac{\nu m^{1/3}}{5\hat{L}b}\leq \frac{\nu m^{1/3}}{2Lb}=\frac{\nu}{2L\eta_0} \leq \frac{\nu}{2L\eta_k}$ for any $k\geq 0$.
 Consider $0 < \eta_k \leq 1$ and $m\geq (cb)^3$, we have $\beta_{k+1} = c\eta_k^2 \leq \frac{cb^2}{m^{2/3}} \leq 1$.
 At the same time, we have $c\leq \frac{m^{2/3}}{b^2}$.
 According to Lemma \ref{lem:C1}, we have
 \begin{align}
 & \frac{1}{\eta_k}\mathbb{E} \|\nabla f(\theta_{k+1}) - u_{k+1}\|^2 -  \frac{1}{\eta_{k-1}}\mathbb{E} \|\nabla f(\theta_k) - u_k\|^2 \nonumber \\
 & \leq \big(\frac{1-\beta_{k+1}}{\eta_k} - \frac{1}{\eta_{k-1}}\big)\mathbb{E} \|\nabla f(\theta_k) -u_k\|^2 + \frac{2\beta^2_{k+1}\sigma^2}{\eta_k} + 4\hat{L}^2\eta_k\|\tilde{\theta}_{k+1}-\theta_k\|^2 \nonumber \\
 & = \big(\frac{1}{\eta_k} - \frac{1}{\eta_{k-1}} - c\eta_k\big)\mathbb{E} \|\nabla f(\theta_k) -u_k\|^2 + 2c^2\eta^3_k\sigma^2 + 4\hat{L}^2\eta_k\|\tilde{\theta}_{k+1}-\theta_k\|^2 \nonumber \\
 & = \big(\frac{1}{b}\big( (m+k)^{\frac{1}{3}} - (m+k-1)^{\frac{1}{3}}\big) - c\eta_k\big)\mathbb{E} \|\nabla f(\theta_k) -u_k\|^2 + 2c^2\eta^3_k\sigma^2 + 4\hat{L}^2\eta_k\|\tilde{\theta}_{k+1}-\theta_k\|^2 \nonumber \\
 & \leq \big(\frac{2}{3b^3}\eta_k - c\eta_k\big)\mathbb{E} \|\nabla f(\theta_k) -u_k\|^2 + 2c^2\eta^3_k\sigma^2 + 4\hat{L}^2\eta_k\|\tilde{\theta}_{k+1}-\theta_k\|^2,
 \end{align}
 where the last inequality holds by
 the following inequality
 \begin{align}
 (m+k)^{\frac{1}{3}} - (m+k-1)^{\frac{1}{3}} & \leq \frac{1}{3(m+k-1)^{2/3}} \leq  \frac{1}{3\big(m/2+k\big)^{2/3}} \nonumber \\
 & \leq \frac{2^{2/3}}{3(m+k)^{2/3}} = \frac{2^{2/3}}{3b^2}\frac{b^2}{(m+k)^{2/3}}= \frac{2^{2/3}}{3b^2}\eta_k^2 \leq \frac{2}{3b^2}\eta_k,
 \end{align}
 where the first inequality holds by the concavity of function $f(x)=x^{1/3}$, \emph{i.e.}, $(x+y)^{1/3}\leq x^{1/3} + \frac{y}{3x^{2/3}}$; the second inequality is due to $m\geq 2$, and
 the last inequality is due to $0<\eta_k\leq 1$. Let $c \geq \frac{2}{3b^3} + \frac{20\hat{L}^2\lambda^2}{\nu^2}$, we have
 \begin{align} \label{eq:F1}
 & \frac{1}{\eta_k}\mathbb{E} \|\nabla f(\theta_{k+1}) - u_{k+1}\|^2 - \frac{1}{\eta_{k-1}}\mathbb{E} \|\nabla f(\theta_k) - u_k\|^2 \nonumber \\
 & \leq -\frac{20\hat{L}^2\lambda^2}{\nu^2}\eta_k\mathbb{E} \|\nabla f(\theta_k) -u_k\|^2 + 2c^2\eta^3_k\sigma^2 + 4\hat{L}^2\eta_k\|\tilde{\theta}_{k+1}-\theta_k\|^2.
 \end{align}
 Here we simultaneously consider $c \geq \frac{2}{3b^3} + \frac{20\hat{L}^2\lambda^2}{\nu^2}$, $c\leq \frac{m^{2/3}}{b^2}$ and $0< \lambda \leq \frac{\nu m^{1/3}}{5\hat{L}b}$,
 we have
 \begin{align}
  \frac{2}{3b^3} + \frac{20\hat{L}^2\lambda^2}{\nu^2} \leq \frac{2}{3b^3} + \frac{20\hat{L}^2}{\nu^2} \frac{\nu^2 m^{2/3}}{25\hat{L}^2b^2} = \frac{2}{3b^3} +  \frac{4m^{2/3}}{5b^2}\leq \frac{m^{2/3}}{b^2}.
 \end{align}
 Then we have $m\geq (\frac{5}{6b})^{2/3}$.

 Next we define a \emph{Lyapunov} function $\Omega_k = \mathbb{E}\big[ f(\theta_k) + \frac{\nu}{16\hat{L}^2\lambda\eta_{k-1}}\|\nabla f(\theta_k)-u_k\|^2\big]$ for any $k\geq 1$.
 According to Lemma \ref{lem:B1}, we have
 \begin{align} \label{eq:F2}
 \Omega_{k+1} - \Omega_k & = f(\theta_{k+1}) - f(\theta_k) + \frac{\nu}{16\hat{L}^2\lambda}\bigg(\frac{1}{\eta_k}\mathbb{E}\|\nabla f(\theta_{k+1})-u_{k+1}\|^2 - \frac{1}{\eta_{k-1}}\mathbb{E} \|\nabla f(\theta_k)-u_k\|^2 \bigg) \nonumber \\
 & \leq \frac{\eta_k\lambda}{\nu}\|\nabla f(\theta_k)-u_k\|^2 -\frac{\nu\eta_k}{2\lambda}\|\tilde{\theta}_{k+1}-\theta_k\|^2-\frac{5\lambda\eta_k}{4\nu}\mathbb{E} \|\nabla f(\theta_k) -u_k\|^2\nonumber \\
 & \quad + \frac{\nu\eta_k}{4\lambda}\mathbb{E}\|\tilde{\theta}_{k+1}-\theta_k\|^2 + \frac{\nu c^2\eta^3_k\sigma^2}{8\hat{L}^2\lambda} \nonumber \\
 & \leq -\frac{\lambda\eta_k}{4\nu}\mathbb{E} \|\nabla f(\theta_k) -u_k\|^2-\frac{\nu\eta_k}{4\lambda}\mathbb{E} \|\tilde{\theta}_{k+1}-\theta_k\|^2 + \frac{\nu c^2\eta^3_k\sigma^2}{8\hat{L}^2\lambda},
 \end{align}
 where the first inequality is due to the above inequality (\ref{eq:F1}).
 Thus, we can obtain
 \begin{align} \label{eq:F3}
 \frac{ \lambda\eta_k}{4\nu}\mathbb{E} \|\nabla f(\theta_k) - u_k\|^2 + \frac{\nu\eta_k}{4\lambda}\mathbb{E}\|\tilde{\theta}_{k+1}-\theta_k\|^2 \leq \Omega_k - \Omega_{k+1} + \frac{\nu c^2\eta^3_k\sigma^2}{8\hat{L}^2 \lambda}.
 \end{align}

 Taking average over $k=1,2,\cdots,K$ on both sides of (\ref{eq:F3}), we have
 \begin{align}
  & \frac{1}{T} \sum_{k=1}^K\mathbb{E} \big[\frac{\lambda\eta_k}{4\nu}\mathbb{E} \|\nabla f(\theta_k) -u_k\|^2 + \frac{\nu\eta_k}{4\lambda}\mathbb{E}\|\tilde{\theta}_{k+1}-\theta_k\|^2 \big] \nonumber \\
  & \leq \frac{f(\theta_1) - f(\theta_{K+1})}{K} + \frac{\nu\|\nabla f(\theta_1) - u_1\|^2}{16\hat{L}^2 \eta_0\lambda K} - \frac{\nu\|\nabla f(\theta_{K+1}) - u_{K+1}\|^2}{16\hat{L}^2\eta_K\lambda K} + \frac{1}{K}\sum_{k=1}^K\frac{\nu c^2\eta^3_k\sigma^2}{8\hat{L}^2 \lambda} \nonumber \\
  & \leq \frac{J(\theta_{K+1})-J(\theta_1)}{K} + \frac{\nu\sigma^2}{16\hat{L}^2\eta_0\lambda K} + \frac{1}{K}\sum_{k=1}^K\frac{\nu c^2\eta^3_k\sigma^2}{8\hat{L}^2\lambda}
  \nonumber \\
  & \leq  \frac{J^*-J(\theta_1)}{K} + \frac{\nu\sigma^2}{16\hat{L}^2 \eta_0\lambda K} + \frac{1}{K}\sum_{k=1}^K\frac{\nu c^2\eta^3_k\sigma^2}{8\hat{L}^2 \lambda},
 \end{align}
 where the second inequality is due to $u_1 = -g(\tau_1|\theta_1)$, $\nabla f(\theta_1)=-\nabla J(\theta_1)$ and Assumption 1,
 and the last inequality holds by Assumption 2.
 Since $\eta_k$ is decreasing, i.e., $\eta_K^{-1} \geq \eta_k^{-1}$ for any $0< k \leq K$, we have
 \begin{align} \label{eq:F4}
  & \frac{1}{K} \sum_{k=1}^K\mathbb{E} \big[ \frac{1}{4\nu^2}\mathbb{E} \|\nabla f(\theta_k) -u_k\|^2 + \frac{1}{4\lambda^2}\mathbb{E}\|\tilde{\theta}_{k+1}-\theta_k\|^2 \big] \nonumber \\
  & \leq  \frac{J^*-J(\theta_1)}{ K\nu\lambda\eta_K} + \frac{\sigma^2}{16\hat{L}^2\eta_K\eta_0\lambda^2 K} + \frac{1}{K\lambda\eta_K}\sum_{k=1}^K\frac{ c^2\eta^3_k\sigma^2}{8\hat{L}^2\lambda} \nonumber \\
  & \leq  \frac{J^*-J(\theta_1)}{K\nu\lambda\eta_K} + \frac{m^{1/3}\sigma^2}{16b\hat{L}^2\lambda^2\eta_KK}
  + \frac{c^2\sigma^2}{8\hat{L}^2\lambda^2 K\eta_K}\int^K_1\frac{b^3}{m+k}dk \nonumber \\
  & \leq \frac{J^*-J(\theta_1)}{K\lambda\eta_K} + \frac{m^{1/3}\sigma^2}{16b\hat{L}^2\lambda^2\eta_KK}
  + \frac{c^2\sigma^2 b^3}{8\hat{L}^2\lambda^2 K\eta_K}\ln(m+K) \nonumber \\
  & = \bigg( \frac{J^*-J(\theta_1)}{b\nu\lambda} + \frac{m^{1/3}\sigma^2}{16b^2\hat{L}^2\lambda^2}
  + \frac{c^2\sigma^2 b^2}{8\hat{L}^2\lambda^2}\bigg) \frac{(m+K)^{1/3}}{K},
 \end{align}
 where the second inequality holds by $\sum_{k=1}^K \eta_k^3 dk \leq \int^K_1 \eta_k^3 dk = b^3\int^K_1(m+k)^{-1}dk$.

 Let $M'=\frac{J^*-J(\theta_1)}{b\nu\lambda} + \frac{m^{1/3}\sigma^2}{16b^2\hat{L}^2\lambda^2}
  + \frac{c^2\sigma^2 b^2}{8\hat{L}^2\lambda^2}$, the above inequality (\ref{eq:F4}) reduces to
 \begin{align}
 \frac{1}{K} \sum_{k=1}^K\mathbb{E} \big[\frac{1}{4\nu^2} \|\nabla f(\theta_k) -u_k\|^2
  + \frac{1}{4\lambda^2} \|\tilde{\theta}_{k+1}-\theta_k\|^2 \big] \leq \frac{M'}{K}(m+K)^{1/3}.
 \end{align}

 According to Jensen's inequality, we have
 \begin{align}
  & \frac{1}{K} \sum_{k=1}^K\mathbb{E} \big[\frac{1}{2\nu} \|\nabla f(\theta_k) -u_k\|
  + \frac{1}{2\lambda} \|\tilde{\theta}_{k+1}-\theta_k\| \big] \nonumber \\
  & \leq \big( \frac{2}{K} \sum_{k=1}^K\mathbb{E} \big[\frac{1}{4\nu^2} \|\nabla f(\theta_k) -u_k\|^2
  + \frac{1}{4\lambda^2} \|\tilde{\theta}_{k+1}-\theta_k\|^2\big]\big)^{1/2} \nonumber \\
  & \leq \frac{\sqrt{2M'}}{K^{1/2}}(m+K)^{1/6} \leq \frac{\sqrt{2M'}m^{1/6}}{K^{1/2}} + \frac{\sqrt{2M'}}{K^{1/3}},
 \end{align}
 where the last inequality is due to the inequality $(a+b)^{1/6} \leq a^{1/6} + b^{1/6}$ for all $a,b\geq 1$.
 Thus we have
 \begin{align}
  \frac{1}{K} \sum_{k=1}^K\mathbb{E} \big[ \frac{1}{\nu}\|\nabla f(\theta_k) -u_k\| + \frac{1}{\lambda} \|\tilde{\theta}_{k+1}-\theta_k\| \big]
  \leq \frac{2\sqrt{2M'}m^{1/6}}{K^{1/2}} + \frac{2\sqrt{2M'}}{K^{1/3}}.
 \end{align}

 Then by using the above inequality (\ref{eq:E7}), we can obtain
  \begin{align}
  \frac{1}{K} \sum_{k=1}^K \mathbb{E} \|\mathcal{B}^{\psi_k}_{\lambda,\langle \nabla f(\theta_k), \theta\rangle}(\theta_k)\|
  \leq \frac{2\sqrt{2M'}m^{1/6}}{K^{1/2}} + \frac{2\sqrt{2M'}}{K^{1/3}}.
 \end{align}

 \end{proof}
\begin{algorithm}[tb]
\caption{ BGPO Algorithm (Actor-Critic Style) }
\label{alg:1-AC}
\begin{algorithmic}[1] %[1] enables line numbers
\STATE {\bfseries Input:}  Total iteration $K$, tuning parameters $\{\lambda,b,m,c\}$ and mirror mappings $\big\{\psi_k\big\}_{k=1}^K$
are $\nu$-strongly convex functions; \\
\STATE {\bfseries Initialize:} $\theta_1 \in \Theta$, ${\theta}^v_1 \in \Theta^v$ and sample a trajectory $\tau_1$ from $p(\tau |\theta_1)$,
and compute $u_1 = -g(\tau_1|\theta_1)$;\\
\FOR{$k = 1, 2, \ldots, K$}
% \tcp{Update Policy Network}
\STATE \textit{\# Update the policy network}
\STATE Update  $\tilde{\theta}_{k+1} = \arg\min_{\theta\in \Theta} \big\{ \langle u_k, \theta\rangle
+ \frac{1}{\lambda} D_{\psi_k}(\theta,\theta_k) \big\}$; \\
\STATE Update  $\theta_{k+1} = \theta_k + \eta_k(\tilde{\theta}_{k+1}-\theta_k)$ with $\eta_k = \frac{b}{(m+k)^{1/2}}$;
\STATE \textit{\# Update the value network}
\STATE Update $\theta^v_{k+1}$ by solving the subproblem (\ref{eq:57});
\STATE \textit{\# Sample a new trajectory and compute policy gradients}
\STATE Sample a trajectory $\tau_{k+1}$ from $p(\tau |\theta_{k+1})$, and compute
$u_{k+1} = -\beta_{k+1} g(\tau_{k+1}|\theta_{k+1}) + (1-\beta_{k+1})u_k$ with $\beta_{k+1} = c\eta_k$; \\
\ENDFOR
\STATE {\bfseries Output:}  $\theta_{\zeta}$ chosen uniformly random from $\{\theta_k\}_{k=1}^{K}$.
\end{algorithmic}
\end{algorithm}
\begin{algorithm}[tb]
\caption{ VR-BGPO Algorithm (Actor-Critic Style) }
\label{alg:2-AC}
\begin{algorithmic}[1] %[1] enables line numbers
\STATE {\bfseries Input:}  Total iteration $K$, tuning parameters $\{\lambda,b,m,c\}$ and mirror mappings $\big\{\psi_k\big\}_{k=1}^K$
are $\nu$-strongly convex functions; \\
\STATE {\bfseries Initialize:}  $\theta_1 \in \Theta$, ${\theta}^v_1 \in \Theta^v$ and sample a trajectory $\tau_1$ from $p(\tau |\theta_1)$, and compute $u_1 =  -g(\tau_1|\theta_1)$;\
\FOR{$k = 1, 2, \ldots, K$}
\STATE \textit{\# Update the policy network}
\STATE Update  $\tilde{\theta}_{k+1} = \arg\min_{\theta\in \Theta} \big\{ \langle u_k, \theta\rangle + \frac{1}{\lambda} D_{\psi_k}(\theta,\theta_k) \big\}$; \\
\STATE Update $\theta_{k+1} = \theta_k + \eta_k(\tilde{\theta}_{k+1}-\theta_k)$ with $\eta_k = \frac{b}{(m+k)^{1/3}}$;
\STATE \textit{\# Update the value network}
\STATE Update $\theta^v_{k+1}$ by solving the subproblem (\ref{eq:57});
\STATE \textit{\# Sample a new trajectory and compute policy gradients}
\STATE Sample a trajectory $\tau_{k+1}$ from $p(\tau |\theta_{k+1})$, and compute $u_{k+1} = -\beta_{k+1} g(\tau_{k+1}|\theta_{k+1}) + (1-\beta_{k+1})\big[u_k - g(\tau_{k+1} | \theta_{k+1})+ w(\tau_{k+1}|\theta_k,\theta_{k+1}) g(\tau_{k+1}|\theta_k)\big]$
with $\beta_{k+1} = c\eta_k^2$; \\
\ENDFOR
\STATE {\bfseries Output:}  $\theta_{\zeta}$ chosen uniformly random from $\{\theta_k\}_{k=1}^{K}$ .
\end{algorithmic}
\end{algorithm}

\section{Actor-Critic Style BGPO and VR-BGPO Algorithms}
In the experiments, we use the advantage-based policy gradient estimator:
\begin{align}
g(\tau|\theta) = \sum_{t=0}^{H-1} \nabla \log\pi_{\theta}(a_t|s_t)\hat{A}^{\pi_{\theta}}(s_t,a_t),
\end{align}
where $\theta (\in \Theta \subseteq \mathbb{R}^{d})$ denotes parameters of the policy network, and $\hat{A}^{\pi_{\theta}}(s,a)$ is an estimator of the advantage function $A^{\pi_{\theta}}(s,a)$. In using advantage-based policy gradient, we also need the state-value function $V^{\pi_{\theta}}(s)$. Here, we use a value network $V_{\theta^v}(s)$ to approximate the state-value function $V^{\pi_{\theta}}(s)$. Specifically, we solve the following problem to obtain the  value network:
\begin{align} \label{eq:57}
\min_{\theta^v \in \Theta^v}\mathcal{L}(\theta^v) :=  \sum_{t=0}^{H-1}\big(V_{\theta^v}(s_t) - \hat{V}^{\pi_{\theta}}(s_t)\big)^2,
\end{align}
where $\theta^v (\in \Theta^v \subseteq \mathbb{R}^{d_v})$ denotes parameters of the value network, and  $\hat{V}^{\pi_{\theta}}(s)$ is an estimator of the state-value function $V^{\pi_{\theta}}(s)$, which is obtained by the GAE~\cite{schulman2015high}. Then we use the GAE to estimate $\hat{A}^{\pi_{\theta}}$ based on value network $V_{\theta^v}$.
We describe the actor-critic style BGPO and VR-BGPO algorithms in Algorithm \ref{alg:1-AC} and Algorithm \ref{alg:2-AC}, respectively.

\section{Detailed Setup of Experimental Environments and Hyper-parameters}
  \label{appendix:B}

 In this section, we provide the detailed setup of experimental environments and hyper-parameters.
 We first provide the detailed setup of our experiments in Tab.~\ref{tab:settings_section6.2_6.3} and Tab.~\ref{tab:settings_section6.4}. We use ADAM optimizer to optimize value functions for all methods and settings, which is a common practice. The importance sampling weight used for VR-BGPO algorithm is clipped within $[0.5, 1.5]$. The momentum term $\beta_k$ is set to be less or equal than one ($\beta_k = \min(\beta_k, 1.0)$ ) through the whole training process.
 % In the experiments, we generate basic stochastic policy gradients based on GAE~\cite{schulman2015high}.

 BGPO and VR-BGPO algorithms involve 4 hyper-parameters $\{\lambda, b, m, c\}$, which may bring additional efforts for hyper-parameter tuning. However, the actual hyper-parameter tuning is not so hard, and we only use one set of $\{b,m,c\}$ for 9 environments. The strategy of hyper-parameter tuning is to separate the four hyper-parameters into two parts. The first part is $\{b,m,c\}$, which mainly decide when the momentum term $\beta_k$ actually affects ($\beta_k< 1.0 $) updates. The second part, $\lambda$, only affects how fast the policy is learning. To further reduce the complexity of hyper-parameter tuning, we always set $m=2$. By grouping hyper-parameters, we only consider $\lambda$ and how $\beta_k$ changes, which largely simplifies the process of hyper-parameter tuning.

 \begin{table*}[t]
     \centering
     % \resizebox{0.99\textwidth}{!}{
     \resizebox{1.0\textwidth}{!}{
     \begin{tabular}{c|c|c|c}
     \hline
          Environments& CartPole-v1 &Acrobat-v1 & MountainCar-v0\\
          \hline
          \hline
          Horizon & 100 & 500  & 500 \\
          Value function Network sizes & $32\times32$  & $32\times32$  & $32\times32$ \\
          Policy network sizes & $8\times8$  & $8\times8$  & $64\times64 $\\
          Number of timesteps & $5\times10^5$ &$5\times10^6$  & $7.5\times10^6$ \\
          Batchsize & $50$ &$100$  & $100$ \\
          VR-BGPO/BGPO $\{b,m,c\}$ & $\{1.5, 2.0, 25\}$ & $\{1.5, 2.0, 25\}$  & $\{1.5, 2.0, 25\}$\\
          BGPO-$l_p$ $\{\lambda_{p=1.5},\lambda_{p=2.0}, \lambda_{p=3.0}\}$ & $\{0.0064, 0.0016, 0.0008\}$ & $\{0.016, 0.004, 0.001\}$  & $\{0.016, 0.004, 0.001\}$
          \\
          BGPO-Diag/VR-BGPO-Diag $\lambda$& $1\times10^{-3}$ & $1\times10^{-3}$  & $1\times10^{-3}$
          \\
          Value function learning rate & $2.5\times10^{-3}$ & $2.5\times10^{-3}$  & $2.5\times10^{-3}$\\
          \hline
     \end{tabular}
     }
     \caption{Setups of environments and hyper-parameters for experiments in section~6.2 and section~6.3. The learning rate of value functions are the same for all methods.}
     \label{tab:settings_section6.2_6.3}
 \end{table*}

 \begin{table*}[t]
     \centering
     \resizebox{1.0\textwidth}{!}{
     \begin{tabular}{c|c|c|c|c|c|c}
     \hline
          Environments& Pendulum-v2 &DoublePendulum-v2 & Walker2d-v2 & Swimmer-v2 & Reacher-v2&HalfCheetah-v2\\
          \hline
          \hline
          Horizon & 500 & 500  & 500 & 500& 500& 500\\
          Value function Network sizes & $32\times32$  & $32\times32$  & $32\times32$ &  $32\times32$& $32\times32$ &  $32\times32$\\
          Policy network sizes & $64\times64$  & $64\times64$  & $64\times64 $ &  $64\times64$& $64\times64 $ &  $64\times64$\\
          Number of timesteps & $5\times10^6$ &$5\times10^6$  & $1\times10^7$ &$1\times10^7$& $1\times10^7$ &$1\times10^7$ \\
          Batchsize & $100$ &$100$  & $100$ &$100$& $100$ &$100$ \\
          VR-BGPO $\{b,m,c\}$ & $\{1.50, 2.0, 25\}$ & $\{1.50, 2.0, 25\}$  & $\{1.50, 2.0, 25\}$ & $\{1.50, 2.0, 25\}$&  $\{1.50, 2.0, 25\}$ & $\{1.50, 2.0, 25\}$\\
          VR-BGPO $\lambda$ & $1\times10^{-2}$ & $1\times10^{-2}$  & $1\times10^{-2}$ & $5\times10^{-4}$  &  $5\times10^{-4}$  &  $5\times10^{-4}$ \\
          TRPO/PPO learning rate & $2.5\times10^{-3}$ & $2.5\times10^{-3}$  & $2.5\times10^{-3}$ & $2.5\times10^{-3}$  &  $2.5\times10^{-3}$  &  $2.5\times10^{-3}$\\
          MDPO learning rate&  $3\times10^{-3}$ & $3\times10^{-3}$  & $3\times10^{-3}$ & $3\times10^{-3}$  &  $3\times10^{-3}$  &  $3\times10^{-3}$\\
          VRMPO learning rate &  $5\times10^{-3}$ & $3\times10^{-4}$  & $1\times10^{-2}$ & $2\times10^{-4}$  &  $5\times10^{-5}$  &  $5\times10^{-5}$\\
          Value function learning rate & $2.5\times10^{-3}$ & $2.5\times10^{-3}$  & $2.5\times10^{-3}$ & $2.5\times10^{-3}$  &  $2.5\times10^{-3}$  &  $2.5\times10^{-3}$\\

          \hline
     \end{tabular}}
     \caption{ Setups of environments and hyper-parameters for experiments in section~6.4. The learning rate of value functions are the same for all methods.}
     \label{tab:settings_section6.4}
 \end{table*}

 \end{appendices}

\end{document}